\documentclass[10pt,letterpaper]{article}

\usepackage{dirtytalk}
\usepackage{amsfonts}

\usepackage{soul}
\soulregister{\cite}{1}
\soulregister{\ref}{1}

\usepackage[title]{appendix}

\makeatletter
\newcommand\subsubsubsection{\@startsection{paragraph}{4}{\z@}%
  {-3.25ex \@plus -1ex \@minus -.2ex}%
  {1.5ex \@plus .2ex}%
  {\normalfont\normalsize\bfseries}}
\makeatother

\usepackage{caption}
\usepackage[utf8]{inputenc}

\usepackage{hhline}
\usepackage{makecell}
\usepackage{subcaption}
\usepackage{fullpage}
\usepackage{times}
\usepackage{array}
\usepackage{ragged2e}
\usepackage{multirow}
\usepackage{fancyhdr, amssymb}
\usepackage[ruled,vlined]{algorithm2e}
\usepackage[margin = 1 in]{geometry}
\usepackage{graphicx}
\usepackage{outlines}
\usepackage{longtable}
\usepackage{amsmath}

\usepackage{url}
\usepackage[english]{babel} 
\usepackage{blindtext}
\usepackage{xltabular}
\usepackage{arydshln} 
\usepackage{wrapfig}
\usepackage{enumitem}
\usepackage{floatrow}
\usepackage{xfrac}
\usepackage[font={footnotesize}, labelfont={bf}, labelsep=space, justification=justified, skip=0pt]{caption}
\usepackage[hang,flushmargin]{footmisc} 
\usepackage{fancyhdr, amssymb, mathtools}
\usepackage{csquotes}
\usepackage{amssymb}
\usepackage{bbm}
\usepackage{mathtools}
\usepackage{perpage}
\usepackage{multicol} 

\usepackage[dvipsnames,table]{xcolor}
\usepackage[
backend=biber,
style=numeric-comp,
sorting=none,
maxnames = 10
]{biblatex}
\usepackage[colorlinks=true, linkcolor=blue, citecolor=blue, filecolor=blue, urlcolor=blue]{hyperref}
\usepackage{cleveref}

\AtBeginEnvironment{appendices}{\crefalias{section}{appendix}}

\makeatletter
\AtBeginDocument{%
  \g@addto@macro\normalsize{%
    \setlength\abovedisplayskip{8pt}
    \setlength\belowdisplayskip{8pt}
    \setlength\abovedisplayshortskip{8pt}
    \setlength\belowdisplayshortskip{8pt}
    \let\orig@setfontsize\@setfontsize
  }%
}
\makeatother

\DeclareMathSizes{10}{10}{7}{6}

\newcommand{\cmt}[1]{} 

\usepackage{booktabs}
\usepackage{graphicx}
\usepackage{float}
\usepackage{multirow}
\usepackage{enumitem}
\usepackage[table]{xcolor} 

\usepackage[T1]{fontenc}
\usepackage{lmodern}

\newcommand{\simgym}{\textit{SimGym}}
\newcommand{\nomem}{\textit{SimGym w/o Memory}}
\newcommand{\intent}{\textit{Intent Only}}
\newcommand{\generic}{\textit{Generic Persona}}

{\title{\fontsize{14}{14}\selectfont 
SimGym: Traffic-Grounded Browser Agents for Offline A/B Testing in E-Commerce
}}

\date{\vspace{-5ex}}
\usepackage{authblk}
\author[1]{Alberto Castelo\thanks{\noindent Corresponding Author: 
alberto.castelo@shopify.com}}
\author[1]{Zahra Zanjani Foumani}
\author[1]{Ailin Fan}
\author[1]{Keat Yang Koay}
\author[1]{Vibhor Malik}
\author[1]{Yuanzheng Zhu}
\author[1]{Han Li}
\author[1]{Meysam Feghhi}
\author[1]{Ronie Uliana}
\author[1]{Shuang Xie}
\author[1]{Zhaoyu Zhang}
\author[1]{Angelo Ocana Martins}
\author[1]{Mingyu Zhao}
\author[1]{Francis Pelland}
\author[1]{Jonathan Faerman}
\author[1]{Nikolas LeBlanc}
\author[1]{Aaron Glazer}
\author[1]{Andrew McNamara}
\author[1]{Lingyun Wang\thanks{\noindent Corresponding Author: lingyun.wang@shopify.com}}
\author[1]{Zhong Wu}

\affil[1]{Shopify, Ottawa, Ontario, Canada}

\begin{document}
    \include{pythonlisting}
    \pagenumbering{arabic}
    \sloppy
    \maketitle
    \begin{abstract} 
A/B testing remains the gold standard for evaluating e-commerce UI changes, yet it diverts traffic, takes weeks to reach significance, and risks harming user experience. We introduce SimGym, a scalable system for rapid offline A/B testing using traffic-grounded synthetic buyers powered by Large Language Model agents operating in a live browser. SimGym extracts per-shop buyer profiles and intents from production interaction data, identifies distinct behavioral archetypes, and simulates cohort-weighted sessions across control and treatment storefronts. We validate SimGym against real human outcomes from real UI changes on a major e-commerce platform under confounder control. Even without alignment post training, SimGym agents achieve state of the art alignment with observed outcome shifts and reduces experiment cycles from weeks to under an hour , enabling rapid experimentation without exposure to real buyers.

\end{abstract}
    \section{Introduction} \label{sec: intro}
A/B testing remains the gold standard for evaluating UI changes in e-commerce, enabling merchants to make data-driven decisions that directly impact conversion rates and revenue. 
However, traditional A/B testing carries significant costs: splitting traffic exposes real users to potentially suboptimal experiences, achieving statistical significance typically requires weeks of data collection, and failed treatments degrade customer experiences before they can be detected and rolled back. These constraints make merchants risk-averse; rather than testing bold redesigns that could dramatically improve or harm conversion, they default to incremental, safe changes.

Solving the aforementioned issues motivates a compelling question: what if \textbf{\textit{synthetic buyers}} could pre-test interface variants before real customers are exposed?
Recent advances in large language model (LLM)-based agents make this increasingly plausible. Autonomous agents can now navigate complex web interfaces \cite{zhou2023webarena,deng2023mind2web,chezelles2024browsergym}, complete multi-step tasks across diverse frontends \cite{pan2024webcanvas,chezelles2024browsergym}, and exhibit realistic decision-making patterns when equipped with persona-driven prompts \cite{park2023generative,wang2023recagent,zhang2024generative}. In the e-commerce domain specifically, systems like PAARS \cite{mansour2025paars} and Shop-R1 \cite{zhang2025shop} have demonstrated that LLM agents can simulate believable shopping behavior when aligned to historical customer data. Most recently, AgentA/B \cite{wang2025agenta} proposed a framework for running synthetic experiments using LLM personas at scale.

Despite these advances, a critical gap remains: no prior work has established predictive validity against real human behavioral outcomes in production e-commerce environments. Existing approaches predominantly validate against task completion metrics, behavioral similarity, or offline benchmarks like WebShop \cite{yao2022webshop} and ShoppingBench \cite{wang2025shoppingbench}. \cite{sun2025llm} validates against clickstreams but not intervention outcomes. AgentA/B emphasizes framework design and behavioral fidelity but does not report causal predictive validity.  In other words, prior works demonstrate that agents can behave like shoppers, but not that they can predict how real shoppers will respond to UI changes. An agent that browses realistically but fails to predict actual conversion shifts provides limited value for pre-testing decisions. Furthermore, most benchmarks evaluate agents on single, standardized interfaces rather than the heterogeneous landscape of real e-commerce, where agents must generalize across thousands of unique themes, layouts, and design patterns. 

\begin{figure*}[t]
  \vskip 0.2in
  \begin{center}
    \centerline{\includegraphics[width=\textwidth]{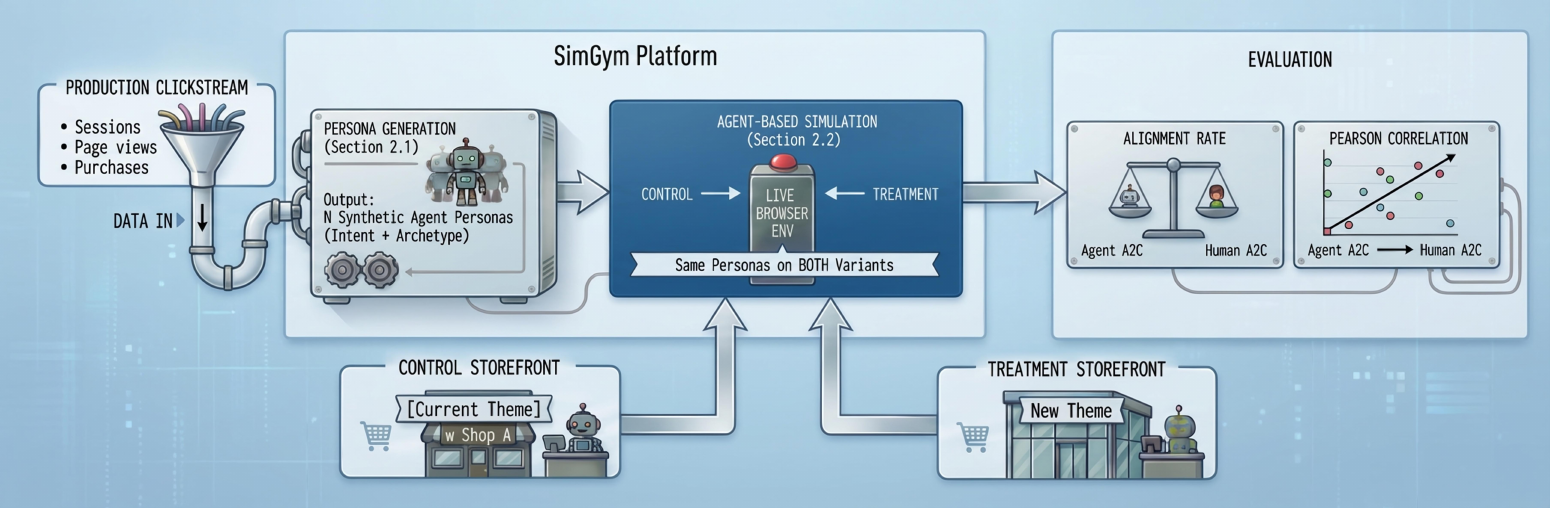}}
    \caption{\simgym~Framework Overview. Production clickstream data feeds into the 
    persona generation pipeline (\Cref{sec: intent_persona_pipeline}), which outputs synthetic agent personas. 
    These agents browse both control and treatment storefronts in a live browser 
    (\Cref{sec: agent_arch}). Evaluation compares agent A2C rates against human outcomes via 
    alignment rate and Pearson correlation (\Cref{sec: eval}).}
    \label{fig:proposed_framework}
  \end{center}
  \vskip -0.2in
\end{figure*}

In this paper, we introduce \simgym, a simulation environment where artificial intelligence (AI) agent shoppers, modeled after a merchant's real customer distribution, autonomously browse two versions of the same online storefront. Unlike offline simulation approaches that operate on static datasets \cite{wang2025shoppingbench,zhang2025shop}, \simgym~agents interact with live storefronts in a full browser environment, requiring real-time perception, memory management, and action execution. By aggregating agents' behaviors and reasonings, \simgym~produces comparative metrics that reveal how UI changes affect add-to-cart (A2C) performance across customer segments. \Cref{fig:proposed_framework}  illustrates our proposed end-to-end pipeline.
Our contributions are as follows:
\begin{itemize}[nosep]
    \item We demonstrate that LLM-based agents can reliably predict A/B test outcomes, enabling faster experimentation cycles and reduced operational costs.
    \item We introduce a novel framework for constructing synthetic user intents and behavioral personas grounded in real-world merchant data, yielding agent behaviors that exhibit strong alignment with empirical customer decision-making patterns.
    \item We validate our approach through deployment on a production platform serving millions of merchants, demonstrating both the scalability and practical efficacy of AI-driven experimentation at scale.
\end{itemize}

The rest of the paper is organized as follows. \Cref{sec: methods} presents our proposed \simgym~platform. We then describe our ground-truth dataset construction and evaluation methodology in \Cref{sec: data_and_eval}. \Cref{sec: results} reports experimental results and ablation studies validating our agents predictive validity. Finally, we conclude the paper and discuss future directions in \Cref{sec: conclusion}.

\textbf{We highlight that this is not the final version of our paper and more analysis will be added per revision.}

    \section{Simgym Platform} \label{sec: methods}
In this section, we first explain our intent and persona generation pipeline in \Cref{sec: intent_persona_pipeline} which transforms raw clickstream data into structured agent profiles. Then, \cref{sec: agent_arch}  presents our proposed agent architecture that executes simulated shopping sessions across control and treatment storefronts.

\subsection{Intent and Persona Generation Pipeline} \label{sec: intent_persona_pipeline}
Building synthetic shoppers that reliably predict real customer behavior requires personas grounded in each merchant's own traffic data. Every storefront attracts a unique distribution of buyers, a boutique fitness brand draws different customers than a general sporting goods retailer. Traditional persona design relies on UX intuition: fictional personas that may not reflect actual behavioral distributions. In this paper, rather than relying on generic personas or transferring personas from similar shops, we extract intents and behavioral profiles directly from each shop's clickstream. Capturing these nuances at scale poses three challenges: $(1)$ personas must reflect the actual distribution of buyer types visiting each merchant's storefront, $(2)$ intents must be calibrated to observed conversion propensities, and $(3)$ the framework must generalize across millions of diverse storefronts without manual tuning.

To address these challenges, we construct agent personas through a six-stage pipeline that transforms raw clickstream data into structured prompts. Each agent comprises two components: a buyer intent specifying the shopping goal and product preferences, and a buyer persona encoding behavioral dimensions (exploration depth, price sensitivity) and value orientations. Our pipeline first clusters sessions to capture purchase patterns (Stage 1), then operates in two parallel tracks: one extracts product preferences and generates intents (Stages 2–3), while the other aggregates buyer-level behavior to construct personas (Stages 4–5). Finally, we compose the agent prompt by combining intent and persona (Stage 6). Our proposed six-stage pipeline, illustrated in \Cref{fig:persona-pipeline}, is as follows:

\textbf{Stage 1: Session-Level Clustering.} 
We represent each storefront session as a feature vector spanning engagement (duration, event count), exploration depth (product views, distinct items), search behavior, funnel progression (A2C, checkout, purchase), and economic value (cart value, order value). Features are standardized via z-scoring across sessions, and we apply k-means clustering with k-means++ initialization for stability. Optimization uses a maximum-iteration cap with early stopping based on centroid displacement. We select $k=5$ by evaluating within-cluster sum of squares, relative changes in inertia, and cluster size balance, alongside qualitative interpretability checks. Each session is assigned to its nearest centroid, and we retain the standardized distance as a confidence score.

\textbf{Stage 2: Product Preference Extraction.} For each (shop, cluster) pair, we compile an input record containing shop metadata (name, industry), the cluster identifier, and an aggregated summary of all browsed and purchased products. An LLM prompted as an e-commerce analyst returns a structured JSON with three fields: $(1)$ product categories: up to ten broad themes, e.g., "sneakers," "athletic wear", $(2)$ individual products: up to ten items unusually popular in this cluster, $(3)$ a brief reasoning. Categories are constrained to generic descriptors to ensure downstream re-usability.

\begin{table*}[t]
  \caption{Buyer Persona Dimensions}
  \label{tab:persona-dimensions}
  \begin{center}
    \begin{small}
        \begin{tabular}{llp{10cm}}
          \toprule
          Dimension & Type & Description \\
          \midrule
          Price sensitivity & Behavioral & Premium / mid-range / budget, inferred from browsed vs.\ purchased price gaps with category-aware normalization \\
          Exploration depth & Behavioral & 0--1 score based on session duration, searches, and product views \\
          Premium focus & Values & Attention to luxury, craftsmanship, prestige \\
          Performance focus & Values & Emphasis on durability, reliability, specifications \\
          Ethics focus & Values & Interest in sustainability, ethical sourcing \\
          \bottomrule
        \end{tabular}
    \end{small}
  \end{center}
  \vskip -0.1in
\end{table*}

\textbf{Stage 3: Buyer Intent Generation.} In this stage we generate structured intents that specify what each agent seeks and whether they intend to purchase or browse.
\begin{itemize}[nosep]
    \item Calibrating the mix: For each shop-cluster, we calibrate the ratio of purchase-focused to browsing-focused intents based on observed behavior. Let $\bar{a}$ denote the cluster's mean A2C and $n$ the number of agents. We compute: $\text{purchase count} = \text{clip}(\text{round}(\bar{a} \cdot n), 1, n-1)$ with the remainder assigned to browsing-focused intents. 
    \item Intent template: Each intent follows a strict two-sentence format: "You are looking for [category]. [Cart intention]." The first sentence names a product category from Stage 2 without qualifiers or specific product names. The second expresses purchase readiness (e.g., "You are ready to purchase") or research mode (e.g., "You are researching options").
    \item Constraints and diversity. To ensure reusability, we forbid mentions of bundles, sizes, discounts, and UI details. Within these constraints, we encourage diversity by spreading intents across the cluster's categories and varying phrasing (e.g., "ready to purchase," "planning to buy," "intending to make a purchase").
\end{itemize}

\textbf{Stage 4: Buyer Behavior Aggregation.} For each (shop, cluster) pair, we select the $n$ sessions closest to the cluster centroid, where $n$ matches the number of agents to deploy. Selecting centroid-proximate sessions ensures we capture the most representative examples of each behavioral cluster. We then aggregate these sessions to the buyer grain, computing cross-session summaries: session counts, funnel rates (A2C, checkout, purchase), average cart and order values, and product interaction histories.

\textbf{Stage 5: Buyer persona Construction.} Rather than characterizing shopping preferences to single, mutually exclusive labels, we adopt a multi-dimensional framework in which multiple traits can coexist. For example, an activity may reflect both budget-conscious and ethics-focused tendencies. We represent shopping behavior in the dataset along five continuous dimensions spanning two categories (see \Cref{tab:persona-dimensions}). Behavioral dimensions describe how shopping unfolds and are derived from aggregated clickstream metrics. We normalize price sensitivity in a category-aware manner, recognizing that $\$30$ may be budget for shoes but premium for a water bottle. Values dimensions capture why products are chosen. For each axis (premium, performance, ethics), we identify products using relevant keywords (e.g., "handcrafted," "organic," "commercial grade") and compute the share of such products among browsed versus purchased items. For instance, if the activity reflects $30\%$ ethical products in browsing but $100\%$ in purchasing, this indicates a stronger revealed preference for sustainability at purchase time. Finally, we prompt an LLM to return a structured JSON containing estimated scores, confidence estimates, and reasoning traces. See \Cref{fig:archetype-extraction} for the full schema and example output.

\textbf{Stage 6: Prompt Composition.} We establish a $1:1$ mapping between intents and personas, then compose each agent prompt by combining the intent's goal (e.g., ``You are looking for running shoes. You are ready to purchase.'') with the persona's behavioral and values profile and product preferences. This yields agents aligned with observed product interests, calibrated to cluster-level conversion propensities, and behaviorally rich with multi-dimensional, overlapping traits. The framework is extensible: additional dimensions (e.g., brand loyalty, social proof responsiveness) can be added without altering the core pipeline. See \Cref{fig: output-persona} as a visualization of our intent and persona extracted output.

\subsection{Agent Architecture} \label{sec: agent_arch}
As noted in \Cref{sec: intro}, unlike offline approaches that operate on static datasets \cite{zhang2025shop}, \simgym~agents interact with live storefronts in a full browser environment. This requires real-time perception, memory management, and robust action execution. Our proposed architectural components to enable agents to operate reliably across the diverse and dynamic landscape of real e-commerce storefronts are (see \Cref{fig:agent-arch} for a visual illustration):

\textbf{Web Perception.} Our agents perceive pages through an accessibility tree, a simplified hierarchical view of page elements (headings, buttons, links) with unique identifiers, rather than raw HTML or screenshots. This approach preserves the semantic structure needed for navigation while filtering irrelevant styling and reducing token consumption \cite{deng2023mind2web,zhou2023webarena}. Each interactive element receives a reference ID mapped to its DOM location, enabling unambiguous action targeting and generalization across diverse storefront themes.

\textbf{Prompt Structure.}  Each step constructs a prompt to provide the LLM with context for decision-making containing: $(i)$ the shopping goal, $(ii)$ buyer persona including exploration depth, price sensitivity, and preferences, $(iii)$ session memory of prior actions and observations, $(iv)$ current page state (URL and accessibility tree), and $(v)$ behavioral constraints governing checkout and cart operations. The model returns a structured response containing reasoning, a termination decision, and the next action. We use schema-constrained generation to ensure well-formed outputs.

\textbf{Step Execution.} Agents follow a perceive-plan-act loop common in autonomous agent architectures \cite{park2023generative}. At each step, the agent first captures the current page state via the accessibility tree to identify available interactive elements. It then constructs a prompt incorporating its goal, buyer persona, session memory, and current state, from which the LLM generates reasoning about goal progress and the next action to take. If the agent determines it should continue, it executes the planned action in the browser and captures the results. The agent then records its reasoning and action outcome to session memory to maintain context across the browsing journey. Finally, it checks termination conditions: the loop continues until the agent reaches its goal, hits a step limit, or triggers a guardrail. We provide a detailed example of an agent's step-by-step reasoning and actions in \Cref{sec:agent-trace}.

\textbf{Memory.} \simgym~agents maintain episodic memory that accumulates throughout the browsing session. Unlike approaches relying on fixed context windows or retrieval-augmented memory \cite{gur2023real}, our agents carry forward a complete session log containing initial navigation context, step-by-step reasoning and decisions, action outcomes, and error states. This full-session memory is passed to the LLM at each step to enable coherent behavior across extended journeys; agents recognize when they have already viewed certain products or attempted certain actions, mirroring how human shoppers remember their browsing context. 

\textbf{Guardrails.} Autonomous agents in open-ended web environments require safeguards against failure modes. \simgym~implements several mechanisms: $(i)$ infinite loop protection: detects repeated identical actions and prevents unbounded looping \cite{pan2024webcanvas}, $(ii)$ step and time limits: enforces bounded execution per simulation, $(iii)$ model retry logic: retries failed LLM calls with error context to improve recovery, and $(iv)$ error propagation: passes browser action failures to the agent for informed recovery rather than blind retry. These guardrails ensure graceful termination when agents encounter unexpected states or reasoning errors.
    \section{Ground Truth Generation and Evaluation}\label{sec: data_and_eval}
Validating that \simgym~agents can predict real customer behavior requires addressing two challenges. First, we must establish a ground-truth dataset that captures how real humans respond to online store changes while controlling for confounding factors. Second, we need evaluation metrics that assess predictive validity across diverse buyer types. We address these challenges in \Cref{sec: data_generation,sec: eval}, respectively.

\subsection{Ground Truth} \label{sec: data_generation}
To validate \simgym's predictive validity, we construct a ground-truth dataset from real UI experiments observed on the e-commerce platform. Each storefront has a single active theme controlling its visual design and layout; when merchants switch themes, we can track customer behavior before and after the change. We select shops based on two criteria: $(1)$ sufficient pre- and post-change traffic for reliable human A2C estimation, and $(2)$ measurable differences in A2C rates between theme versions. However, a simple pre-post comparison is insufficient to attribute behavioral shifts to the theme change itself, as other factors such as seasonal effects, promotional periods, new store ramp-up, or pricing changes could also affect the difference. We therefore filter our observations to include only shops free from these confounders, and validate this filtering approach using double machine learning \cite{chernozhukov2018double} to confirm consistent treatment effect estimates.

Since theme changes vary in magnitude, from minor styling tweaks to complete layout redesigns, we employ an LLM evaluator using both visual screenshots and parsed DOM to characterize each change and stratify the dataset by the impact of the theme change. We aggressively filter theme transitions to reduce confounding factors (promotions, seasonality, merchandising changes, ramp-up periods, pricing or assortment shifts) and to ensure changes are non-trivial and observable via DOM tree parsing. This yields a final dataset of $20$ shops, each containing the theme transition, a summary of what changed, and observed human behavioral shifts (A2C, conversion) between themes. Future benchmark versions will expand as we incorporate vision-enabled agents, allowing inclusion of transitions whose effects are primarily visual.

\subsection{Evaluation} \label{sec: eval}
We evaluate \simgym~along two dimensions: predictive validity against real human outcomes, and agent behavioral quality.
For predictive validity, we measure whether agents predict the direction and magnitude of behavioral changes rather than mimicking exact browsing sequences. We use A2C rate as our primary metric since it directly indicates purchase intent and aligns with what merchants want to optimize. We define the alignment rate as the percentage of shops where the direction of human A2C change between themes matches the direction of agent A2C change. To incorporate agent uncertainty in the directional comparison, we also define an alignment probability. For each shop, we assign a continuous value in $[0,1]$ representing the probability that the agent's directional change agrees with the human directional change based on the Bayesian posterior probability of the agent’s A2C direction change. Then we use Pearson correlation coefficient to capture how well the magnitude of agent-predicted changes correlates with observed human changes within each buyer cluster.

To ensure metrics are not inflated by spurious behavior, an LLM evaluator reviews simulation logs and flags undesirable outcomes (agents stuck in loops, out-of-stock attempts, intent mismatches). Flagged sessions are then examined during analysis to understand what's driving the alignment and correlation metrics, helping ensure they reflect genuine predictive capability rather than agent failures.
    
\section{Experiments and Results}
\label{sec: results}
\begin{figure}[ht]
  \vskip 0.1in
  \begin{center}
    \centerline{\includegraphics[width=0.8\columnwidth]{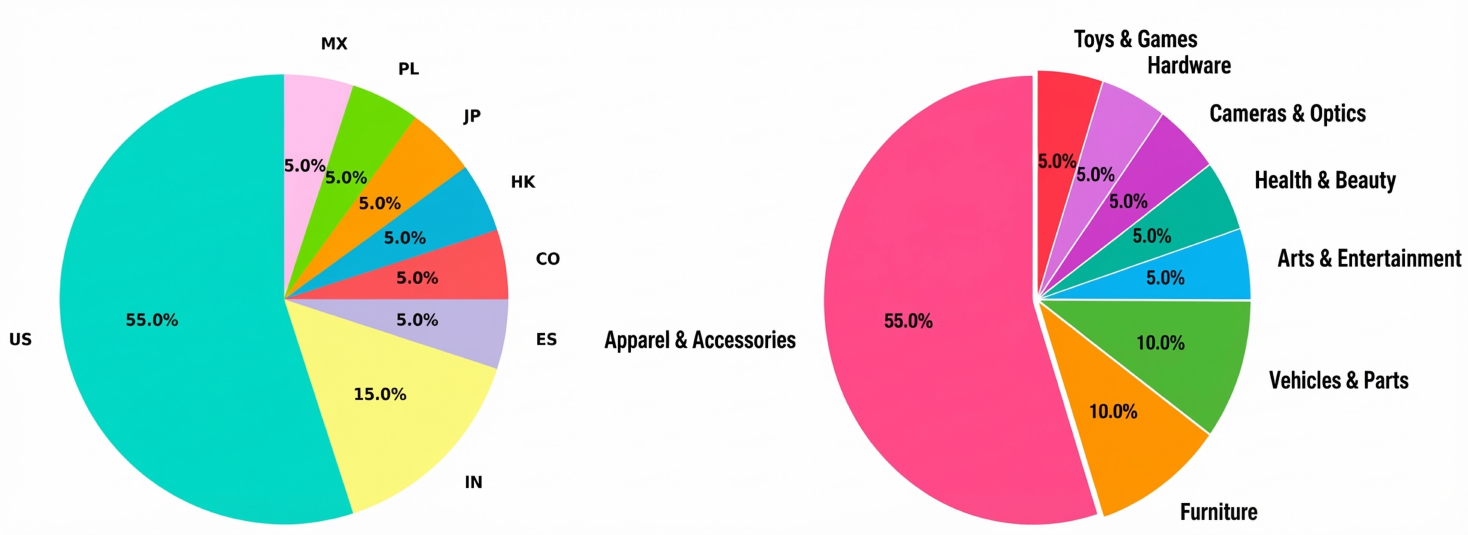}}
    \vskip 0.1in
    \caption{Dataset Distribution Across $20$ Storefronts. Left: Geographic distribution by country. Right: Product category mix across retail domains.} 
    \label{fig:dataset}
  \end{center}
\end{figure}

In this section, we evaluate \simgym's predictive validity on a ground-truth dataset of $20$ shops spanning $12$ countries and diverse industries (\Cref{fig:dataset}), constructed using the method described in \Cref{sec: data_generation}. We start by explaining our agent sample size selection (\Cref{sec: sample-size}), then conduct ablation studies to isolate how each component contributes to predictive validity (\Cref{sec: results}), using the evaluation metrics defined in \Cref{sec: eval}. To ensure robustness throughout our simulations, we repeated all simulations twice and report average values.

\subsection{Agent Sample Size Selection}
\label{sec: sample-size}
\begin{figure}[h]
  \vskip 0.1in
  \begin{center}
    \begin{subfigure}[b]{0.48\columnwidth}
      \centering
      \includegraphics[width=\textwidth]{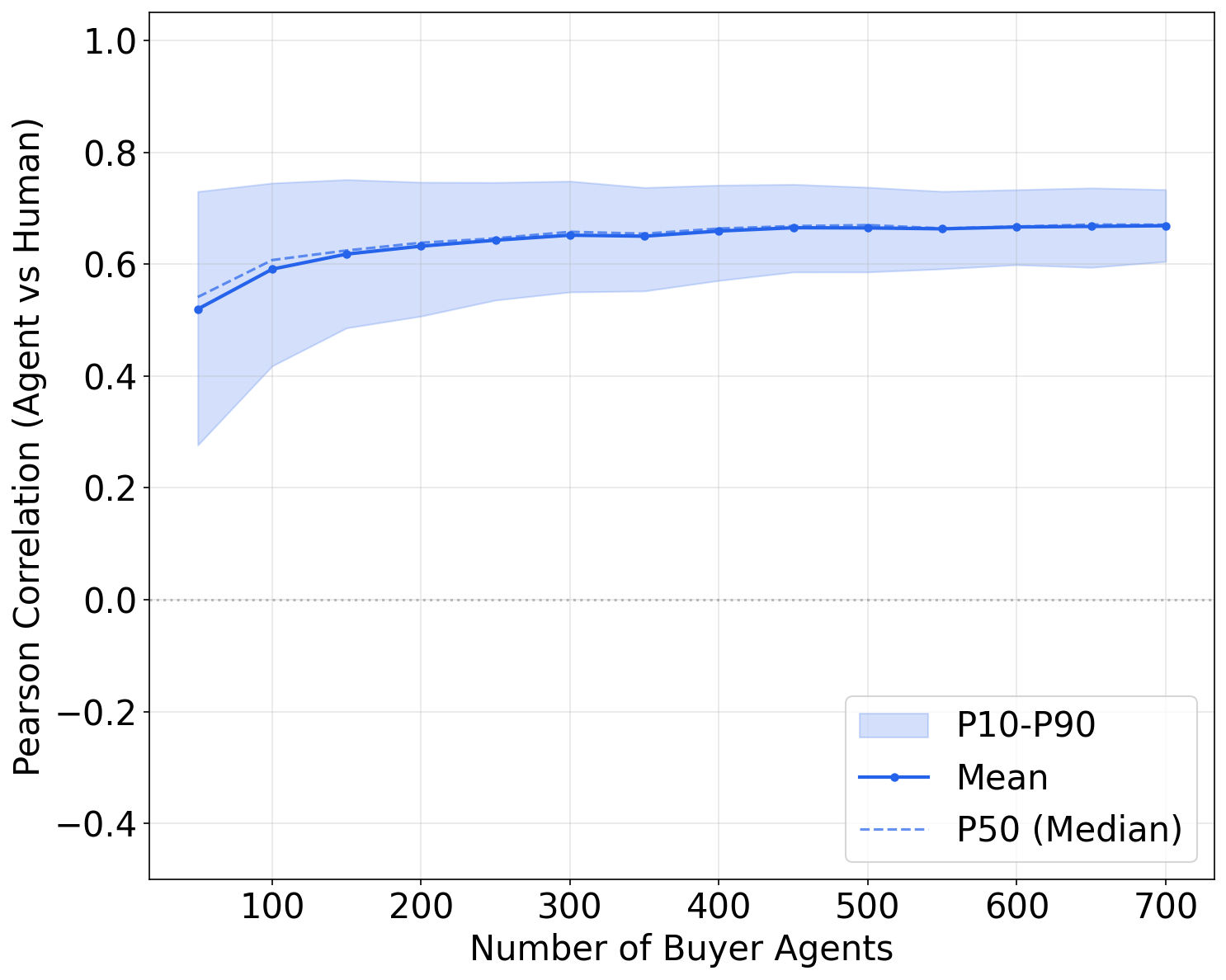}
      \caption{Correlation}
      \label{fig:bootstrap_correlation}
    \end{subfigure}
    \hfill
    \begin{subfigure}[b]{0.48\columnwidth}
      \centering
      \includegraphics[width=\textwidth]{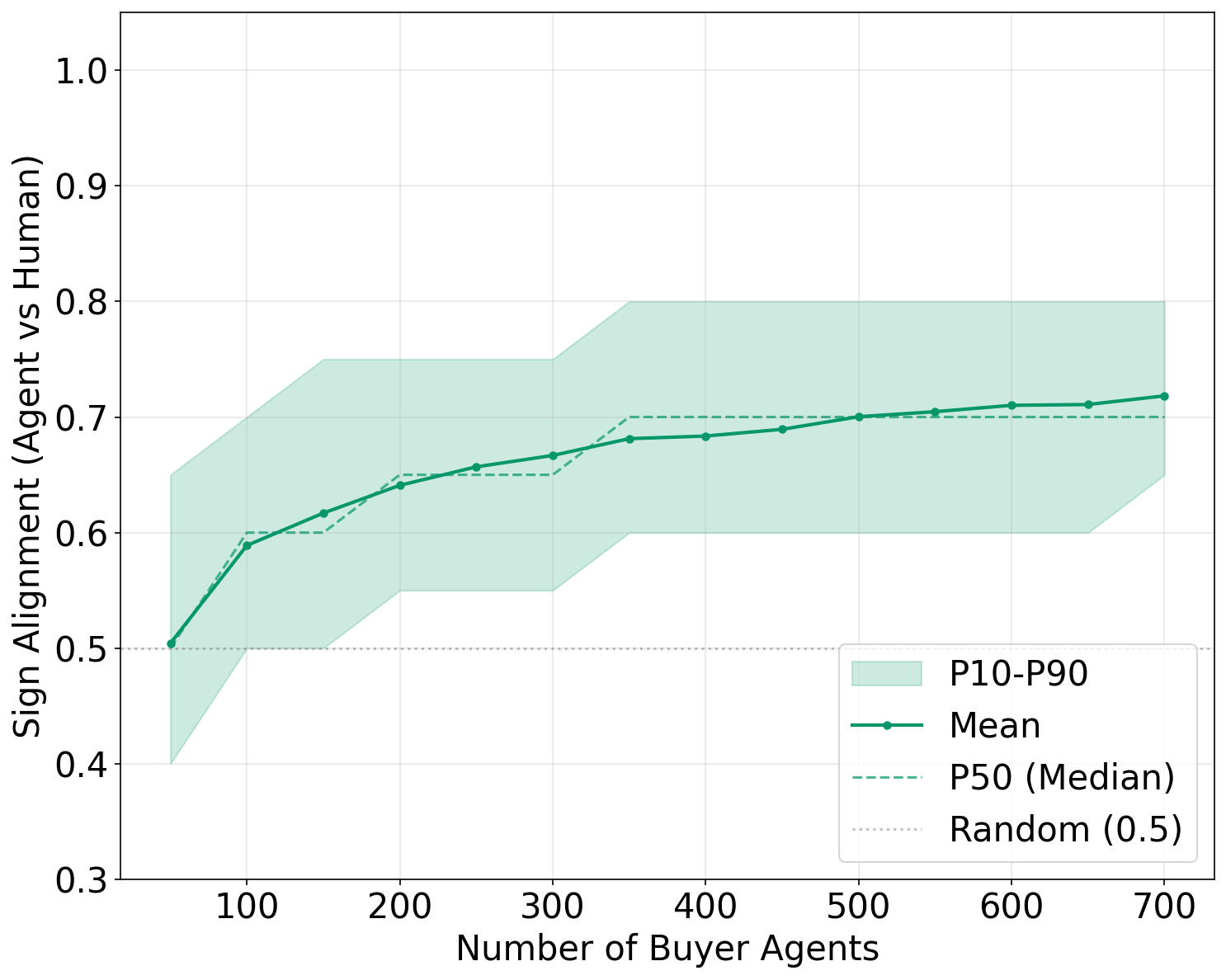}
      \caption{Sign Alignment}
      \label{fig:bootstrap_alignment}
    \end{subfigure}
    \vskip 0.1in
    \caption{Effects of Agent Sample Size on Evaluation Metrics.}
    \label{fig:bootstrap}
  \end{center}
\end{figure}
To determine the number of agents needed for reliable estimates, we conduct a bootstrap resampling analysis. For sample sizes ranging from $50$ to $700$ agents, we sample with replacement from two independent simulation runs with identical settings, evaluate the self-consistency of A2C rate changes across runs, and repeat this process $1000$ times to compute confidence intervals. The solid and dashed lines in \Cref{fig:bootstrap} show the mean and median of the sampling, respectively, with shaded bands indicating the 10th–90th percentile range (P10–P90). The close agreement between mean and median across all sample sizes confirms that the bootstrap distributions are symmetric and well-behaved.

\Cref{fig:bootstrap} shows that both metrics improve with sample size but exhibit diminishing returns. Sign alignment (\Cref{fig:bootstrap_alignment}), which is the percentage of shops where agent A2C changes share the same direction, rises from $\approx51\%$ at $50$ agents to $\approx73\%$ at $700$ agents, with the steepest gains occurring before $400$ agents. Pearson correlation (\Cref{fig:bootstrap_correlation}) stabilizes around $0.65$ after approximately $300$ agents, with the confidence band narrowing considerably. Beyond $500$ agents, both metrics plateau: means stabilize, percentile bands cease narrowing, and additional agents yield marginal improvement while increasing computational cost. 
Based on this analysis, we select $600$ agents per shop; slightly above the plateau threshold to provide margin for session-level failures while maintaining computational feasibility.

\subsection{Ablation Studies}
\label{sec: results}
\simgym's two core components, detailed in \Cref{sec: methods}, each comprise multiple design decisions that collectively enable predictive validity. To isolate the effects of these components on \simgym's alignment with real human outcomes, we ablate one key element from each: episodic session memory from the agent architecture (\Cref{sec: agent_arch}), and per-shop persona grounding from the intent and persona extraction pipeline (\Cref{sec: intent_persona_pipeline}). These ablations target the mechanisms most directly responsible for behavioral 
coherence and buyer-distribution fidelity; two properties we hypothesize are necessary for predicting real conversion shifts. 

We highlight that beyond aggregate metrics, we also analyze the behavioral distribution of agents whose A2C outcomes differed between the pre- and post-theme-change conditions. These distributions are computed over the subset of agents that exhibited different behavior across theme versions, approximately 10–20\% of the $1200$ agents per method.

\subsubsection{Memory Ablation}
To quantify the effects of episodic memory, we compare \simgym~with and without session memory across $20$ storefronts. In the no-memory condition, called \nomem, agents receive only the current page state and their persona at each step, without access to prior actions or observations.

\begin{table}[t]
  \caption{Impact of Session Memory on Predictive Validity}
  \label{tab: memory_ablation}
  \vskip 0.1in
  \begin{center}
        \begin{tabular}{lccc}
          \toprule
          \multirow{2}{*}{Configuration} & \multirow{2}{*}{Correlation} & Alignment & Alignment \\
           & & (\%) & Prob \\
          \midrule
          \simgym & \textbf{0.64} & \textbf{69} & \textbf{0.69} \\
        \textit{w/o Memory}  & 0.29 & 55 & 0.55 \\
          \bottomrule
        \end{tabular}
  \end{center}
  \vskip -0.1in
\end{table}
As shown in \Cref{tab: memory_ablation}, removing memory collapses predictive validity to near-random chance ($69\%$ to $55\%$). The alignment probability is particularly revealing: \simgym's $0.69$ alignment probability indicates confident directional predictions, while memoryless agents' low probability ($0.55$, barely above pure uncertainty) shows that even correct predictions carry little confidence when based on isolated page states rather than coherent browsing journeys. 
To further illustrate the root causes of this degradation, we analyze session-level behaviors.

\begin{table}[h]
  \caption{Memory Ablation Session-Level Outcomes}
  \label{tab: session-outcomes}
  \begin{center}
        \begin{tabular}{lcc}
          \toprule
          Metric & \simgym & \nomem \\
          \midrule
          Goal Reached & 90\% & 45\% \\
          Timeout (Steps Limit) & 9.59\% & 54.93\% \\
          Avg Journey Length & 11.7 steps & 13.7 steps \\
          Journey Length Std Dev & 9.2 steps & 10.7 steps \\
          \bottomrule
        \end{tabular}
  \end{center}
  \vskip -0.1in
\end{table}

As illustrated in \Cref{tab: session-outcomes}, memory substantially impacts session completion. Without session context, over half of agents ($54.93\%$) hit the step limit before reaching a decision (a $5.7 \times$ increase over \simgym), indicating they wander inefficiently rather than progressing toward their goal. Goal completion drops from $90\%$ to $45\%$, and journey length increases with higher variance, confirming that memoryless agents navigate erratically, often revisiting pages or repeating actions they cannot remember having tried.

\begin{table}[b]
  \caption{Memory Ablation Behavioral Distribution}
  \label{tab:failure-modes}
  \begin{center}
        \begin{tabular}{lcc}
          \toprule
          Behavioral Mode      & \simgym  & \nomem \\
          \midrule
          Product Not Found & 36.36\% & 11.11\%   \\
          No A2C Decision   & 40.27\% & 26.67\%   \\
          Failed to Add     & 5.28\%  & 1.04\%    \\
          Stuck in Loop     & 8.50\%  & 45.70\%   \\
          Price Rejection   & 0.88\%  & 0.89\%    \\
          Theme Exit        & 8.70\%  & 14.59\%   \\
          \bottomrule
        \end{tabular}
  \end{center}
  \vskip -0.1in
\end{table}

The behavioral distribution of agents in \Cref{tab:failure-modes}, further reveals why memory matters for predictive validity. With memory, behavioral shifts reflect typical shopping outcomes: agents determine no suitable product exists ($36.36\%$) or reach no A2C decision based on persona preferences ($40.27\%$). Without memory, navigational dysfunction dominates: $45.70\%$ of agents become stuck in loops (vs. $8.50\%$ for \simgym), unable to recognize they have already attempted certain actions. This contrast explains the correlation gap. With memory, agents respond systematically to UI changes and are anchoring the history of what it has observed to their endowed persona and intent; their A2C decisions reflect whether a theme change helps or hurts product discovery. Without memory, misalignment stems from internal incoherence rather than genuine responses to the storefront, producing noise that correlates with chance rather than human outcomes.

\subsubsection{Persona Ablation}
To validate the importance of our proposed traffic-grounded personalization (\Cref{sec: intent_persona_pipeline}), in this section we compare three persona configurations across all $20$ storefronts: \simgym, \intent, and \generic. In all configurations, agents receive shopping intents (e.g., "looking for running shoes"); the difference lies in whether and how behavioral personas are provided.

\intent~baseline removes personas entirely: agents receive shopping goals but no behavioral profile describing how they shop or why they choose products. \generic~tests whether donor-based personas can substitute for shop-specific extraction. To this end, we bucket shops by industry, region, price tier, and catalog size, then within each sub-segment, pool behavioral data from the top $10$ highest-session donor shops. We cluster these pooled sessions and extract persona distributions that represent typical buyer mixes for structurally similar shops. For a target shop, we match it to its sub-segment, sample personas from the pooled priors, and ground intents in the target's catalog categories; ensuring relevance despite using donor behavior. This approach avoids expensive per-shop donor matching while producing more realistic buyer distributions than uniform sampling, since priors reflect actual purchasing patterns within similar shops.

As shown in \Cref{tab:persona}, \intent~achieves only $52\%$ alignment (random chance) and $0.27$ correlation.The behavioral distribution shown in \Cref{tab:persona-ablation-behaivoral}, illuminates the underlying causes of this random behavior. 
To interpret these patterns, we first note from \Cref{tab:persona} that \simgym~ achieves $69\%$ alignment and $0.64$ correlation, demonstrating that its predictions are consistent with real human behavior. This means \simgym's behavioral patterns reflect meaningful, UI-informed decisions rather than noise.
Among these patterns, the critical behavior is ``No A2C Decision''; agents that reach products but choose not to purchase. This is precisely where UI design influences conversion: image quality, layout clarity, and trust signals affect whether customers proceed to buy. Because \simgym~accurately predicts real outcomes, its agents reaching this stage ($40.27\% $of cases) are detecting and responding to the UI changes that drive actual conversion shifts.
\intent~agents reach this UI-sensitive stage in only $24.84\%$ of cases. 

\begin{table}[t]
  \caption{Impact of Persona on Predictive Validity}
  \label{tab:persona}
  \begin{center}
        \begin{tabular}{lccc}
          \toprule
          \multirow{2}{*}{Configuration} & \multirow{2}{*}{Correlation} & Alignment & Alignment \\
           & & (\%) & Prob \\
          \midrule
          \simgym & \textbf{0.64} & \textbf{69} & \textbf{0.69} \\
          \intent & 0.27 & 52 & 0.54 \\
          \generic & 0.27 & 62 & 0.59 \\
          \bottomrule
        \end{tabular}
  \end{center}
  \vskip -0.1in
\end{table}

\begin{table}[h]
  \caption{Persona Ablation Behavioral Distribution}
  \label{tab:persona-ablation-behaivoral}
  \begin{center}
      \begin{tabular}{lccc}
        \toprule
        Behavioral Mode          & \simgym  & \intent & \shortstack{\textit{Generic}\\ \textit{Persona}} \\
        \midrule
        Product Not Found & 36.36\% & 34.99\%     & 43.56\%         \\
        No A2C Decision   & 40.27\% & 24.84\%     & 35.31\%         \\
        Failed to Add     & 5.28\%  & 13.25\%     & 2.23\%          \\
        Other             & 18.90\% & 26.92\%     & 18.81\%         \\
        \bottomrule
      \end{tabular}
  \end{center}
  \vskip -0.1in
\end{table}

Without personas, agents lack browsing strategy. A goal like ``looking for running shoes'' specifies what to seek but not how: whether to explore deeply or skim, which price range to consider, what attributes to prioritize. Without this guidance, agents exhibit operational failures: lower completion rates, higher timeouts (\Cref{tab:session-outcomes-persona}), and the highest ``Failed to Add'' rate (\Cref{tab:persona-ablation-behaivoral}). \intent~outcomes thus depend on whether agents stumble through navigation successfully, not on purchase evaluation. The non-zero correlation exists only because some UI changes happen to affect navigation, but navigation alone cannot capture the product presentation and trust signals that drive real purchasing behavior. The low alignment probability ($0.54$ in \Cref{tab:persona}) further reinforces this interpretation: the alignment probability barely above pure uncertainty, confirms that correct predictions reflect luck rather than systematic response to UI changes.

\begin{table}[t]
  \caption{Persona Ablation Session-Level Outcomes}
  \label{tab:session-outcomes-persona}
  \begin{center}
        \begin{tabular}{lcc}
          \toprule
          Configuration & Goal Reached & Timeout \\
          \midrule
          \simgym & 90\% & 9.59\% \\
          \intent & 79\% & 17.03\% \\
          \generic & 90\% & 9.22\% \\
          \bottomrule
        \end{tabular}
  \end{center}
  \vskip -0.1in
\end{table}

Returning to \Cref{tab:persona}, Generic Persona achieves $62\%$ alignment and $0.27$ correlation, better than Intent Only while worse than \simgym~where fully customized persona is used. The navigation capability is shown in \Cref{tab:session-outcomes-persona}: Generic Persona navigates efficiently ($90\%$ goal reached, $9.22\%$ timeout), comparable to \simgym~and better than Intent Only. 

The behavioral distribution in \Cref{tab:persona-ablation-behaivoral} reveals why \generic~underperforms compared to \simgym. \generic~agents show the highest ``Product Not Found'' rate ($43.56\%$) compared to \simgym~($36.36\%$) and \intent~($34.99\%$). Since navigation succeeds (\Cref{tab:session-outcomes-persona}), the issue is not how agents search but \textit{who} is searching.
Custom personas reflect the target shop's actual buyer distribution, while generic personas rely on segment-level averages that may misrepresent the true audience. For instance, a hardware store's actual customers may skew toward performance-oriented buyers, but its generic persona pool, drawn from broader segment averages, might over-represent premium-seeking agents who reject products lacking luxury signals. This distribution shift means agents search for products that exist but don't match their donor-derived preferences, driving the elevated ``Product Not Found'' rate and preventing agents from reaching the UI-sensitive decision stage where theme design matters.

Noteworthy is to mention that \generic~reaches ``No A2C Decision'' more often than \intent~($35.31\%$ vs. $24.84\%$), and correspondingly, achieves slightly higher predictive validity. This matches the previous observation: \generic~agents are better equipped to decide if a product fits buyer needs, whereas \intent~agents lack persona guidance. However, because \generic~preferences are inferred from donor shops rather than the target shop's true customer base, their reasons for deciding not to purchase may be less representative of what real customers care about, resulting in a lower predictive validity compared to the fully customized persona in \simgym. 
The alignment probability ($0.59$, \Cref{tab:session-outcomes-persona}) also confirms this interpretation: being above $0.50$ indicates predictions are not merely noisy, but systematically better than chance.

\begin{figure}[t]
  \vskip 0.2in
  \begin{center}
    \begin{subfigure}{0.48\columnwidth}
      \centering
      \includegraphics[width=\columnwidth]{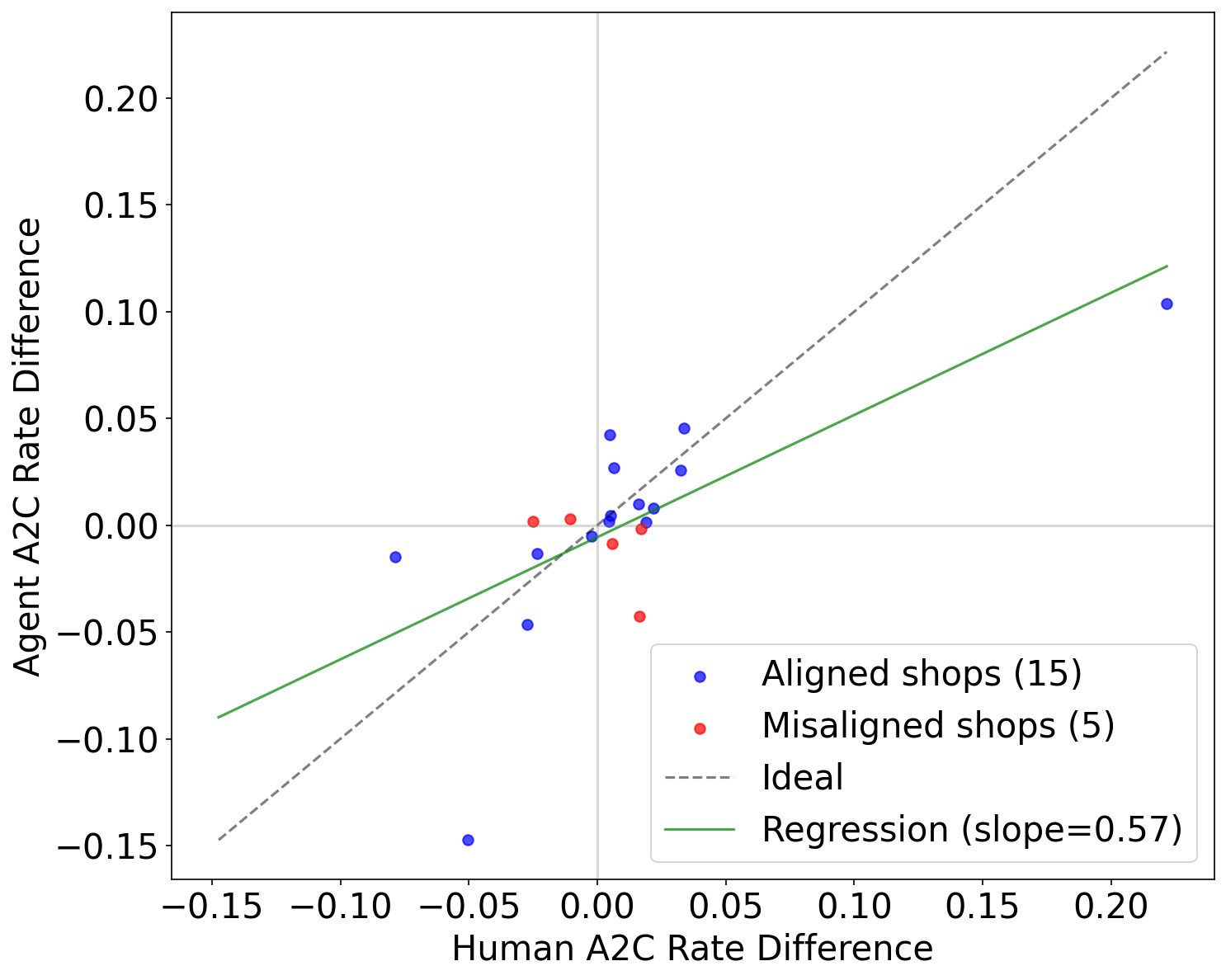}
      \caption{\simgym}
      \label{fig:simgym}
    \end{subfigure}
    \hfill
    \begin{subfigure}{0.48\columnwidth}
      \centering
      \includegraphics[width=\columnwidth]{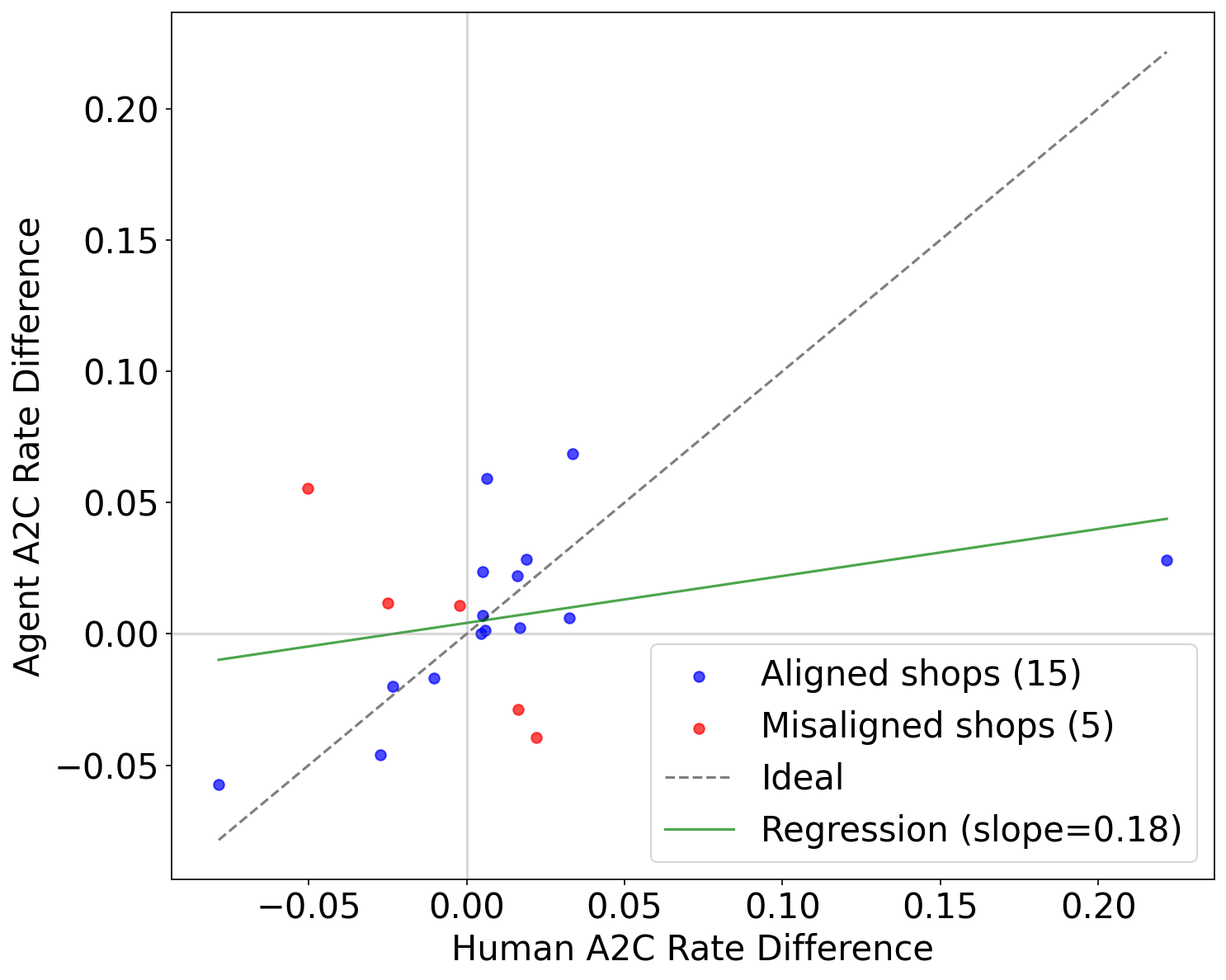}
      \caption{\generic}
      \label{fig:generic-persona}
    \end{subfigure}
    \vspace{0.5em}
    \begin{subfigure}{0.48\columnwidth}
      \centering
      \includegraphics[width=\columnwidth]{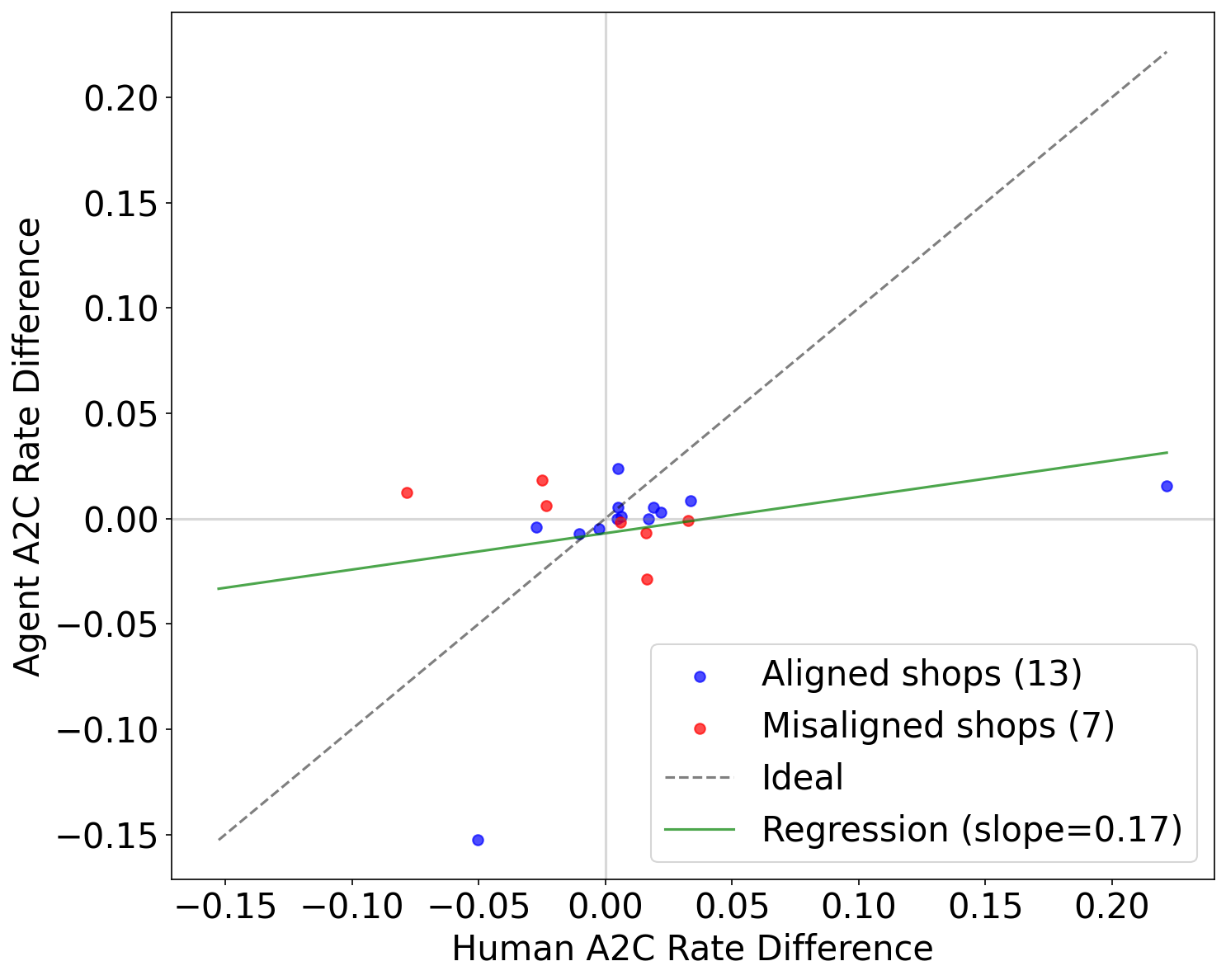}
      \caption{\intent}
      \label{fig:intent-only}
    \end{subfigure}
    \hfill
    \begin{subfigure}{0.48\columnwidth}
      \centering
      \includegraphics[width=\columnwidth]{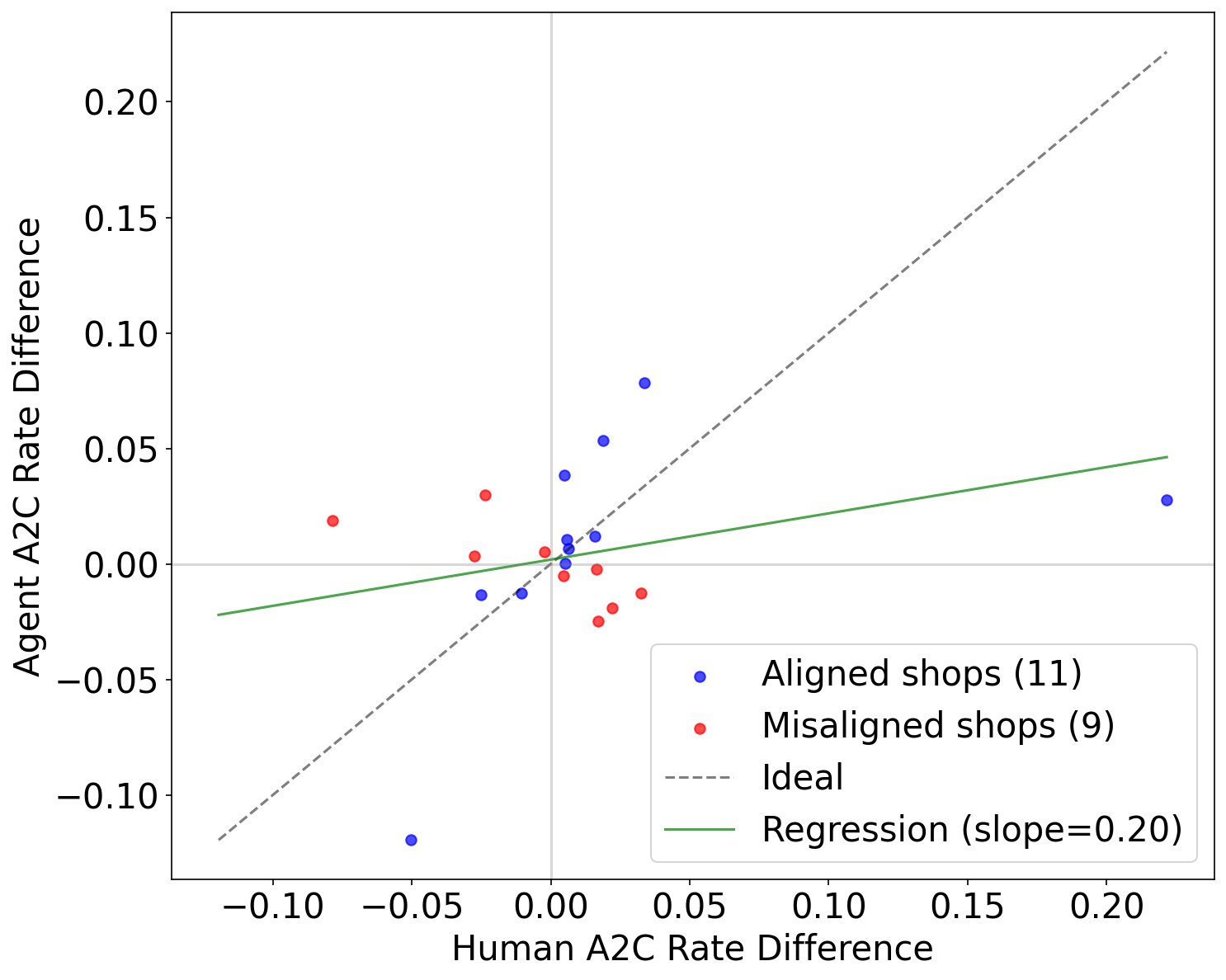}
      \caption{\nomem}
      \label{fig:no-memory}
    \end{subfigure}
    \caption{Predictive Validity Across Configurations.}
    \label{fig:four-panel}
  \end{center}
  \vskip -0.1in
\end{figure}

These ablations converge on a unified finding, visualized in \Cref{fig:four-panel}. The scatter plots display the best-performing simulation for each configuration to illustrate the clearest separation between conditions, while \Cref{tab: memory_ablation,tab:persona} report average values across the two simulations.
Starting with \simgym(\Cref{fig:simgym}), points cluster along the trend line (slope = $0.57$) with $15$ of $20$ shops showing directional alignment by falling in the upper-right or lower-left quadrants, where agent and human behavioral shifts share the same sign. This indicates that agents predict both direction and relative magnitude of behavioral shifts.

The ablated configurations present stark contrasts. \generic~(\Cref{fig:generic-persona}) achieves the same directional alignment as \simgym~($15$ of $20$ shops) but with a dramatically flatter slope ($0.18$ vs. $0.57$). These agents predict the direction of behavioral shifts but not their magnitude; donor-derived preferences provide general guidance, but only shop-specific personas capture how strongly customers respond to UI changes. \intent~(\Cref{fig:intent-only}) shows a weak trend (slope = $0.17$) with $13$ of $20$ shops aligned, yet points remain clustered near zero. This behavior reflects agents that struggle to reach the evaluation stage where UI effects manifest. \nomem~(\Cref{fig:no-memory}) exhibits a similarly shallow slope ($0.20$) with only $11$ of $20$ shops aligned, as agents get stuck in loops rather than responding systematically to UI changes.
The visual contrast between \simgym's tight diagonal clustering and the scattered or compressed distributions of all ablations underscores that reliable prediction requires both components: data-grounded personas to evaluate with relevant preferences and capture effect magnitudes, and memory to navigate coherently to the decision stage.

    \section{Conclusion and Future Work} \label{sec: conclusion}
In this paper, we introduced \simgym, a scalable system for rapid offline A/B testing using traffic-grounded synthetic buyers powered by LLM agents operating in a live browser. Our approach addresses three key challenges in synthetic buyer simulation. First, we generate buyer persona that matches the actual buyer type distribution merchants using production clickstream data. Second, we model buyers along multi-dimensional personas, capturing both how they shop and the characteristics of their product preference. Third, agents operate in live browser environments with episodic memory and guardrails, supporting coherent behavior across extended shopping journeys that are consistent with actual human behavior. Experiments on $20$ shops demonstrate \simgym's high correlation with human A2C changes , reducing experiment cycles from weeks to under an hour while eliminating exposure risk to real buyers. 

Several directions could extend this work. Supervised fine-tuning on human shopping traces or reinforcement learning from behavioral feedback could further improve fidelity. End-to-end learning of personas directly from raw session data may capture subtler behavioral patterns that explicit feature engineering misses. Metrics measuring agents' proficiency in mimicking expert shopping behaviors would enable more granular evaluation beyond A2C alignment. Finally, with agents capable of reliable A/B evaluation, future systems could use SimGym as the inner loop in an automated optimization process: iteratively proposing UI changes, evaluating them in simulation, and surfacing the most promising modifications to merchants for implementation.
    \nocite{*} 
    \printbibliography
    \pagebreak
    \appendix
\begin{multicols}{2}
\section{Simgym Platform}
\subsection{Intent and Persona Generation}
\begin{center}
  \includegraphics[width=0.8\columnwidth]{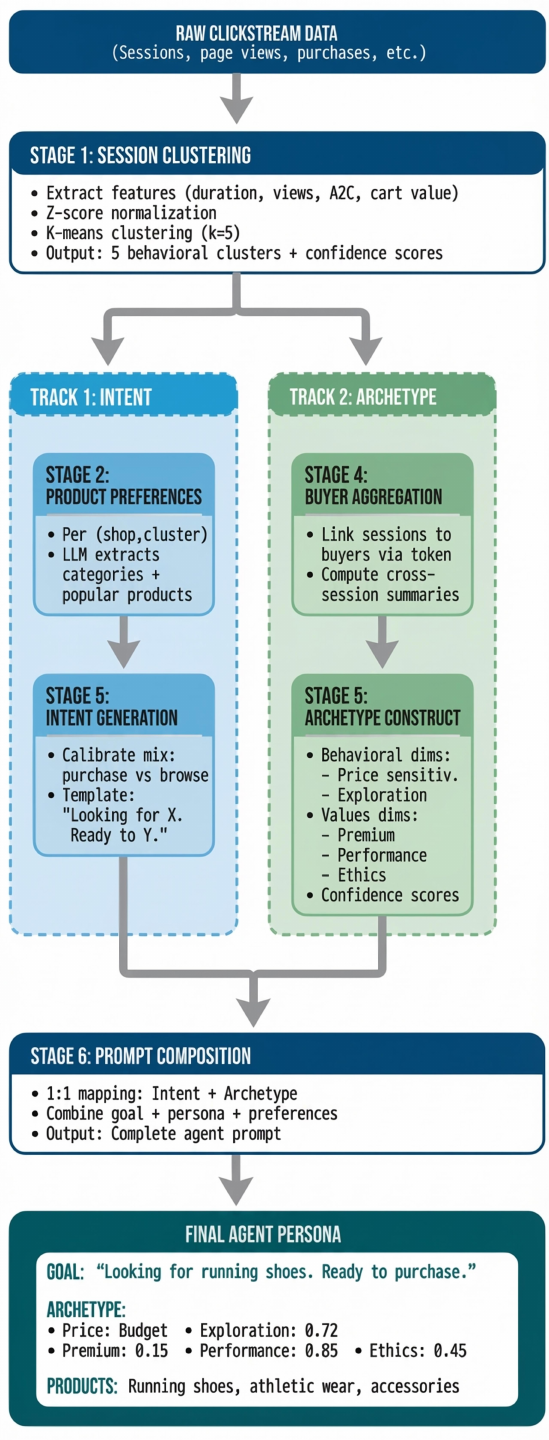}
  \captionof{figure}{Visualization of the Persona Generation Pipeline.}
  \label{fig:persona-pipeline}
\end{center}

\Cref{fig:persona-pipeline} provides a detailed visualization of the six-stage persona generation pipeline summarized in \Cref{sec: intent_persona_pipeline}. 
The pipeline transforms raw clickstream data into structured agent prompts through two parallel tracks: Track 1 generates buyer intents and product preferences (Stages 2-3), while Track 2 constructs multi-dimensional buyer archetypes (Stages 4-5). Both tracks consume data from the initial session clustering (Stage 1) and merge in the final prompt composition (Stage 6).

\subsection{Buyer Archetype Construction}
\label{appendix:archetype}
\begin{center}
  \includegraphics[width=0.7\columnwidth]{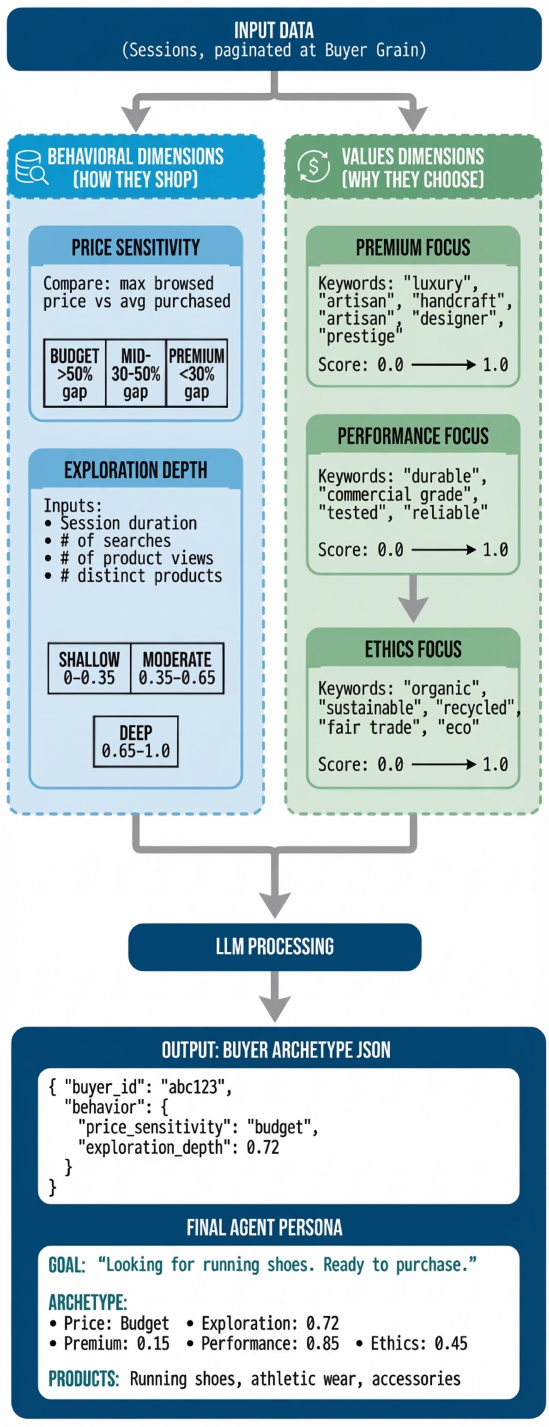}
  \captionof{figure}{Buyer Archetype Construction Framework.}
  \label{fig:archetype-extraction}
\end{center}
\Cref{fig:archetype-extraction} details the buyer archetype construction process 
(Stage 5 of \Cref{fig:persona-pipeline}). Given aggregated buyer data from 
Stage 4, we compute scores along five continuous dimensions organized into two 
categories:

\textbf{Behavioral dimensions} capture \textit{how} buyers shop:
\begin{itemize}[nosep]
    \item \textit{Price sensitivity}: Measures the gap between maximum browsed 
    price and average purchased price. Buyers are labeled as budget ($>$50\% gap), 
    mid-range (30--50\% gap), or premium ($<$30\% gap), with category-aware 
    normalization to account for product-specific price distributions.
    \item \textit{Exploration depth}: A 0--1 score derived from session duration, 
    search count, and product views, mapped to shallow (0--0.35), moderate 
    (0.35--0.65), or deep (0.65--1.0) exploration regimes.
\end{itemize}

\textbf{Values dimensions} capture \textit{why} buyers choose products:
\begin{itemize}[nosep]
    \item \textit{Premium focus}: Attention to luxury, craftsmanship, and prestige.
    \item \textit{Performance focus}: Emphasis on durability, reliability, and specifications.
    \item \textit{Ethics focus}: Interest in sustainability and ethical sourcing.
\end{itemize}

For each values dimension, we identify products containing relevant keywords 
(e.g., ``handcrafted,'' ``commercial grade,'' ``organic'') and compare the 
proportion in browsed vs.\ purchased items to infer revealed preferences. 
An LLM processes these signals with category-aware taxonomy and deterministic 
decision rules, outputting a structured JSON with scores, confidence estimates, 
and reasoning traces.

\subsection{Intent and Persona Extracted Output}
\label{appendix:intent-output}
\begin{center}
  \includegraphics[width=0.8\columnwidth]{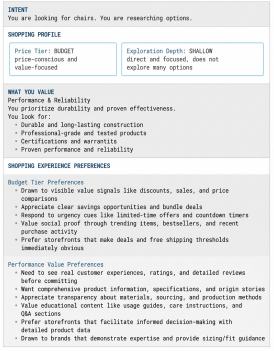}
  \captionof{figure}{Intent and Persona Extracted Output.}
  \label{fig: output-persona}
\end{center}

\Cref{fig: output-persona} presents a representative output from our intent and persona generation pipeline described in \Cref{sec: intent_persona_pipeline}. The visualization illustrates how the six-stage pipeline transforms raw clickstream data into a structured agent prompt.

The Intent component, generated in Stage 3, follows our two-sentence template: the first sentence specifies the product category ("chairs"), while the second sentence indicates browsing mode ("researching options").
The Shopping Profile encodes two behavioral dimensions derived from Stage 4's buyer behavior aggregation and Stage 5's persona construction: $(1)$ Price Tier is set to "Budget" (price-conscious and value-focused), reflecting category-aware price sensitivity computed from the gap between browsed and purchased price points; $(2)$ Exploration Depth is classified as "Shallow" (direct and focused), derived from the buyer's session duration, search frequency, and product view counts.

The Values section captures the buyer's "Performance \& Reliability" orientation, one of three values dimensions in our framework (see \Cref{tab:persona-dimensions}). This classification emerges from Stage 5's analysis of product interaction patterns related keywords (e.g., "durable," "professional-grade," "certified").

Finally, the Shopping Experience Preferences operationalize the persona into actionable behavioral guidance. These preferences are derived from the intersection of behavioral and values dimensions: Budget Tier Preferences specify responsiveness to discount signals, social proof, and urgency cues, while Performance Value Preferences emphasize attention to detailed specifications, customer reviews, and transparency about materials. Together, these components enable the agent to exhibit coherent, persona-consistent behavior throughout the shopping session.

\subsection{Agent Architecture}
\label{appendix:agent-arch}
\begin{center}
  \includegraphics[width=0.8\columnwidth]{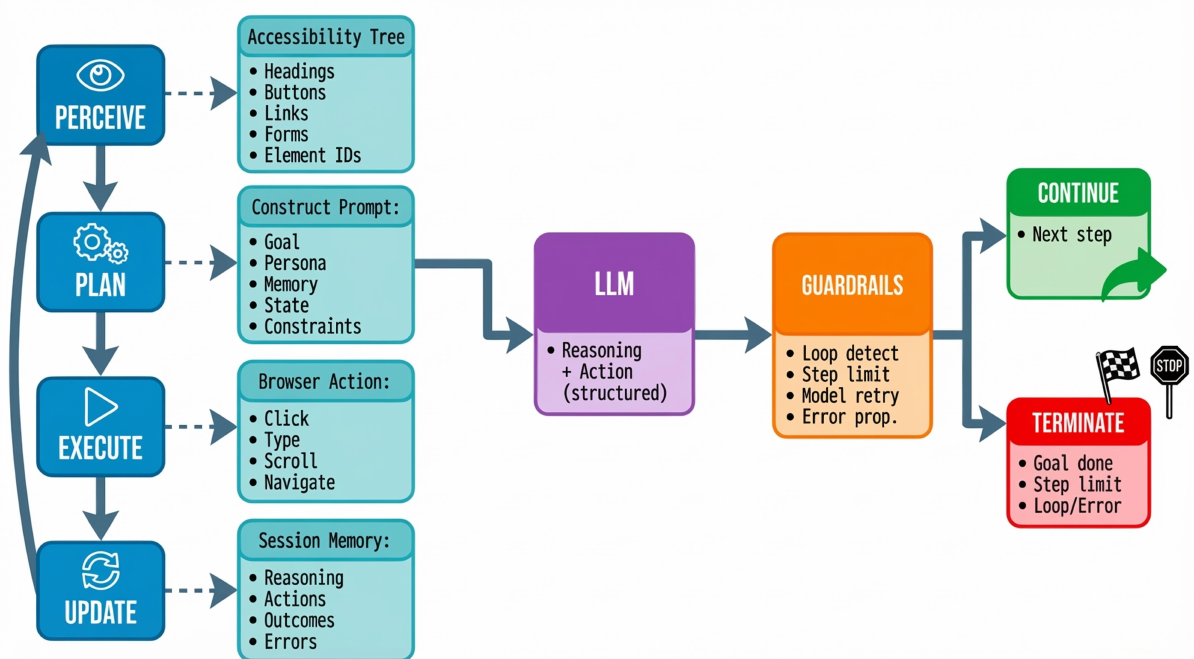}
  \captionof{figure}{Agent Architecture.}
  \label{fig:agent-arch}
\end{center}
\Cref{fig:agent-arch} provides a detailed visualization of the agent 
architecture described in \Cref{sec:agent_arch}. Each agent step follows a 
perceive-plan-execute-update loop:
\begin{itemize}[nosep]
    \item \textbf{Perceive}: Captures the current page via an accessibility tree 
    containing headings, buttons, links, forms, and element IDs.
    \item \textbf{Plan}: Constructs a prompt with goal, persona, memory, current 
    state, and constraints. The LLM generates structured reasoning and a proposed action.
    \item \textbf{Execute}: Performs browser actions (click, type, scroll, navigate) 
    on the live storefront.
    \item \textbf{Update}: Records reasoning, actions, outcomes, and errors to 
    session memory.
\end{itemize}
After each step, guardrails check for failure conditions (loop detection, step/time 
limits, errors). If checks pass, the agent continues to the next step; otherwise, 
it terminates. This loop repeats until the agent completes its goal, reaches a 
limit, or encounters an unrecoverable error.

\subsection{Agent Trace Example}
\label{sec:agent-trace}
\begin{center}
  \includegraphics[width=\columnwidth]{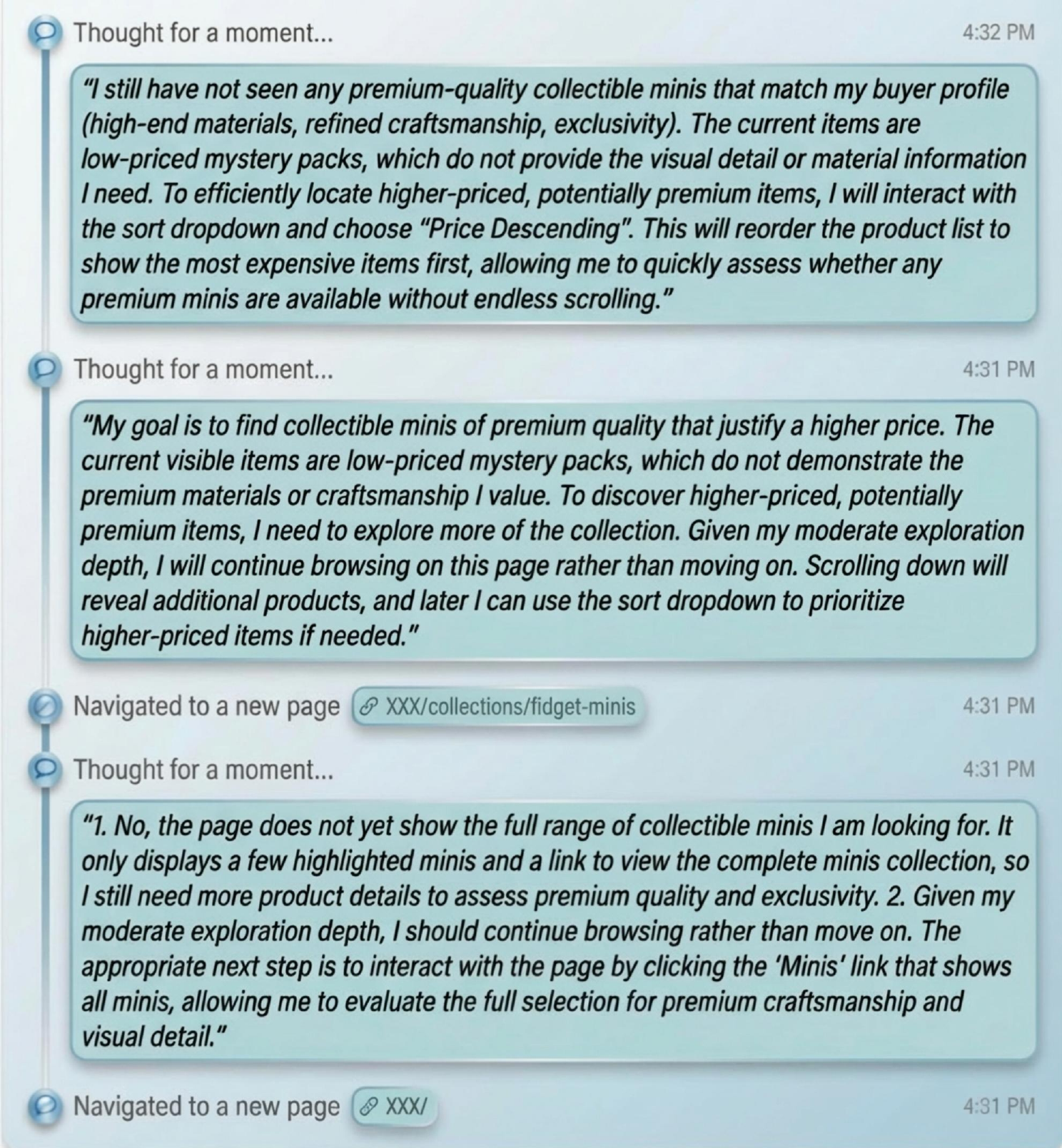}
  \captionof{figure}{Agent Reasoning During Initial Navigation.}
  \label{fig:trace-navigation}
\end{center}
In this section, we present a complete trace of a \simgym~agent completing a shopping task on an anonymous store, specializing in $3D$ printed fidget toys and collectible figures. The agent's goal is to find and purchase a premium-quality collectible mini, with a buyer persona emphasizing luxury materials, refined craftsmanship, and moderate exploration depth.

\Cref{fig:trace-navigation} shows the agent's initial reasoning as it begins navigating the store. The agent explores multiple collections (\Cref{fig:collections}), systematically rejecting low-priced items that don't match its premium preferences. \Cref{fig:trace-exploration} captures the agent's reasoning as it moves between collections: \textit{``The current collection does not contain any high-priced, premium-quality collectible minis that match the buyer profile's emphasis on luxury materials and refined craftsmanship.''} Rather than settling for budget options priced at \$4–7, the agent efficiently shifts between collections; behavior consistent with its moderate exploration depth.
\begin{center}
  \begin{minipage}{0.9\columnwidth}
    \centering
    \includegraphics[width=\textwidth, trim=8 20 20 20, clip]{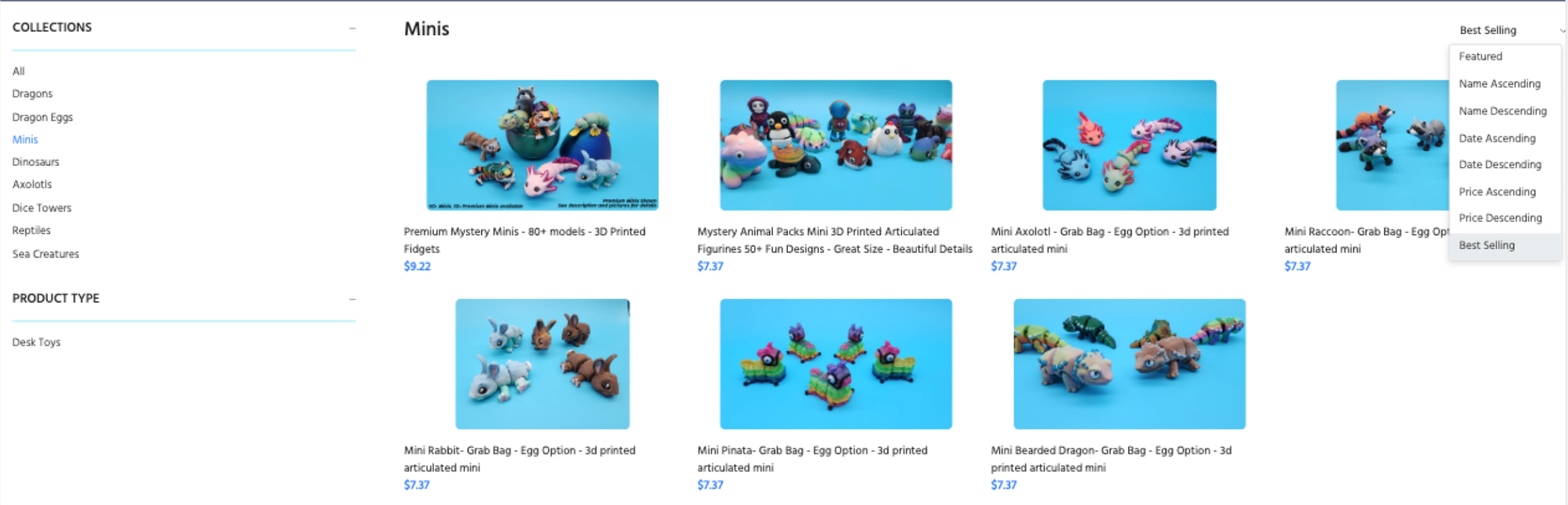}
    \centerline{\small (a) Minis collection (\$7–9).}
    \label{fig:minis-collection}
  \end{minipage}
  
  \vspace{0.5em}
  
  \begin{minipage}{0.9\columnwidth}
    \centering
    \includegraphics[width=\textwidth, trim=8 20 20 20, clip]{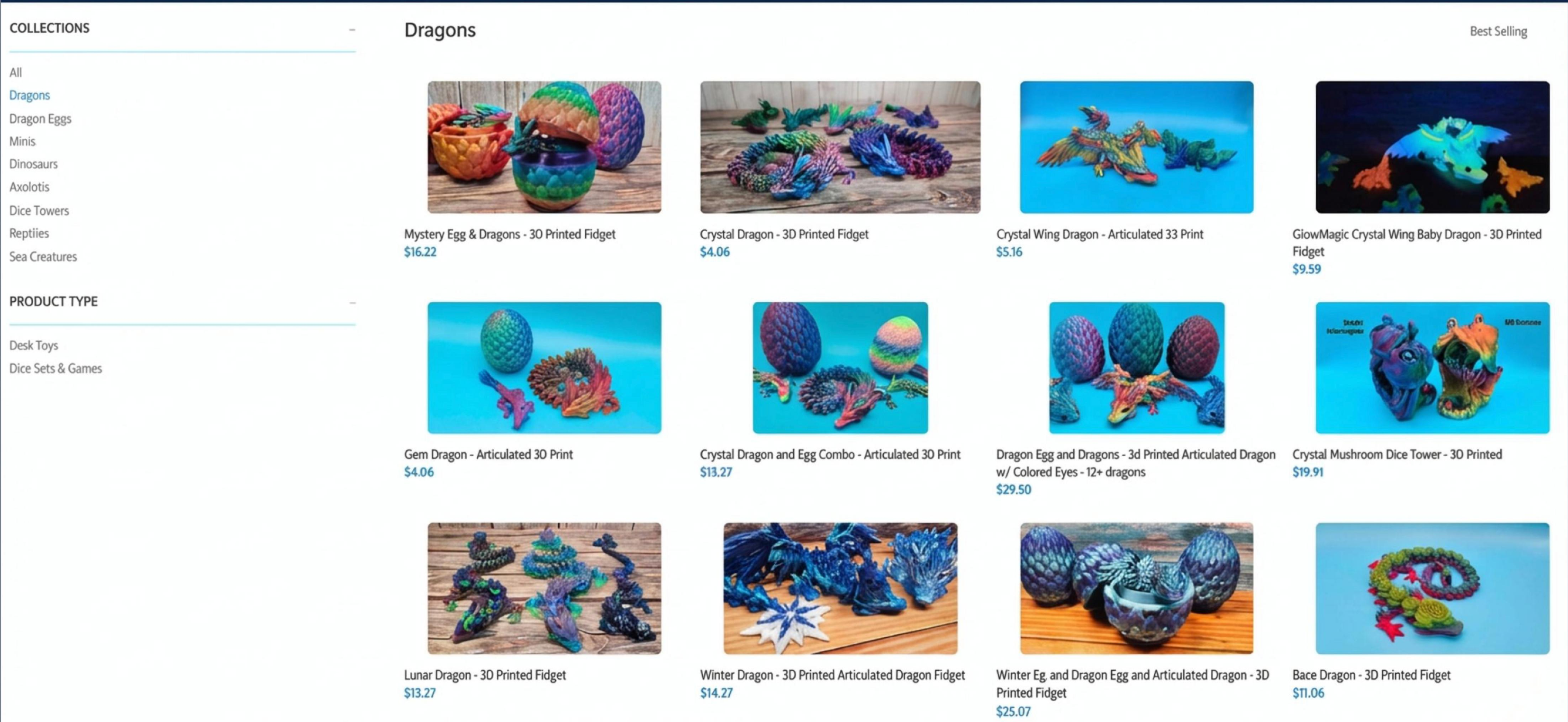}
    \centerline{\small (b) Dragons collection (\$4–\$30).}
    \label{fig:dragons-collection}
  \end{minipage}
  
  \vspace{0.5em}
  
  \begin{minipage}{0.9\columnwidth}
    \centering
    \includegraphics[width=\textwidth, trim=0 20 20 20, clip]{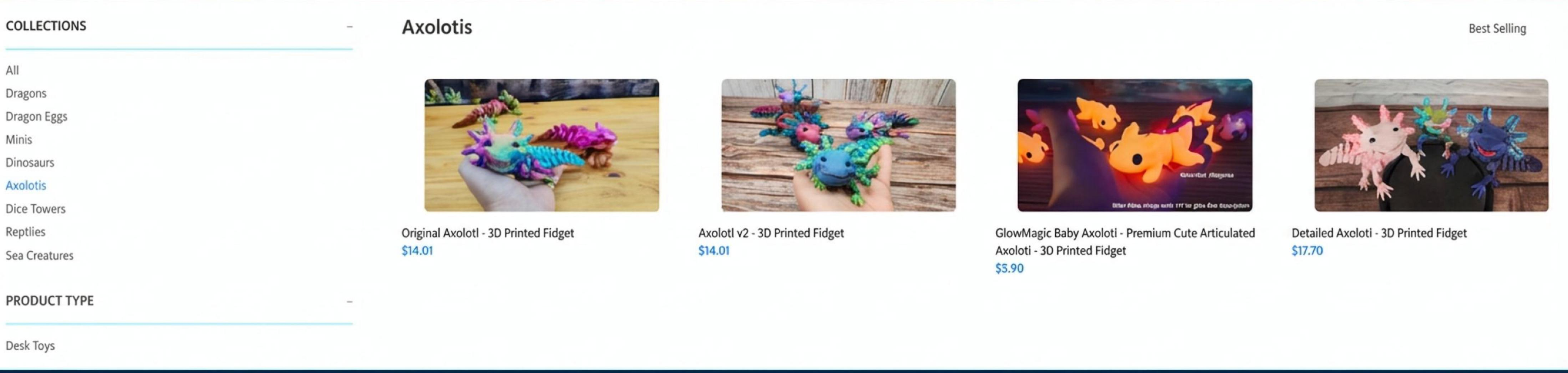}
    \centerline{\small (c) Axolotls collection (\$6–\$18).}
    \label{fig:axolotls-collection}
  \end{minipage}
  
  \captionof{figure}{Screenshots of Agent Browsing.}
  \label{fig:collections}
\end{center}

After exploring multiple collections, the agent identifies a suitable product: the Premium Crystal Wing Dragon (\Cref{fig:product-cart}), priced at \$33.18. The agent's reasoning confirms the match: \textit{``I have found a product that matches my buyer profile: it is presented as a premium edition, uses high-quality 3D printing, offers refined craftsmanship, and is priced higher than the low-cost mystery packs.''} The agent selects product options and adds the item to cart (\Cref{fig:product-cart}). \Cref{fig:trace-checkout} shows the final reasoning as the agent proceeds to checkout, concluding: \textit{``Given the moderate exploration depth, further browsing is unnecessary; proceeding to checkout will fulfill the buying intent.''}
\begin{center}
  \includegraphics[width=0.95\columnwidth]{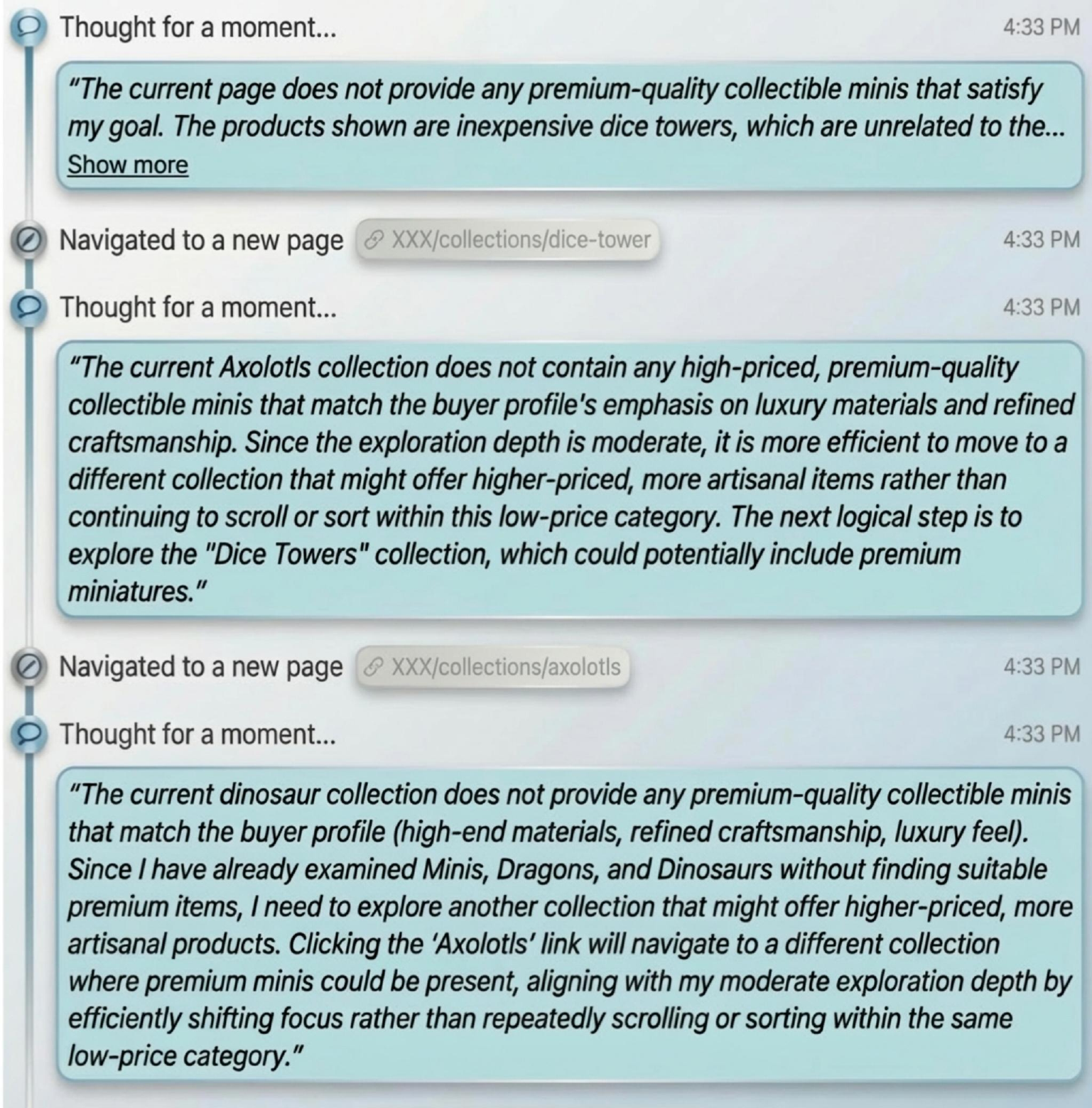}
  \captionof{figure}{Agent Reasoning During Collection Exploration.}
  \label{fig:trace-exploration}
\end{center}

This trace illustrates persona-consistent behavior throughout an extended shopping journey. The agent's premium preference led it to reject multiple budget-priced collections before identifying a suitable \$33.18 product, while its moderate exploration depth prevented excessive browsing once a match was found.

\begin{center}
\begin{minipage}{\columnwidth}
  \centering
  
  \includegraphics[width=0.9\columnwidth, trim=8 0 0 0, clip]{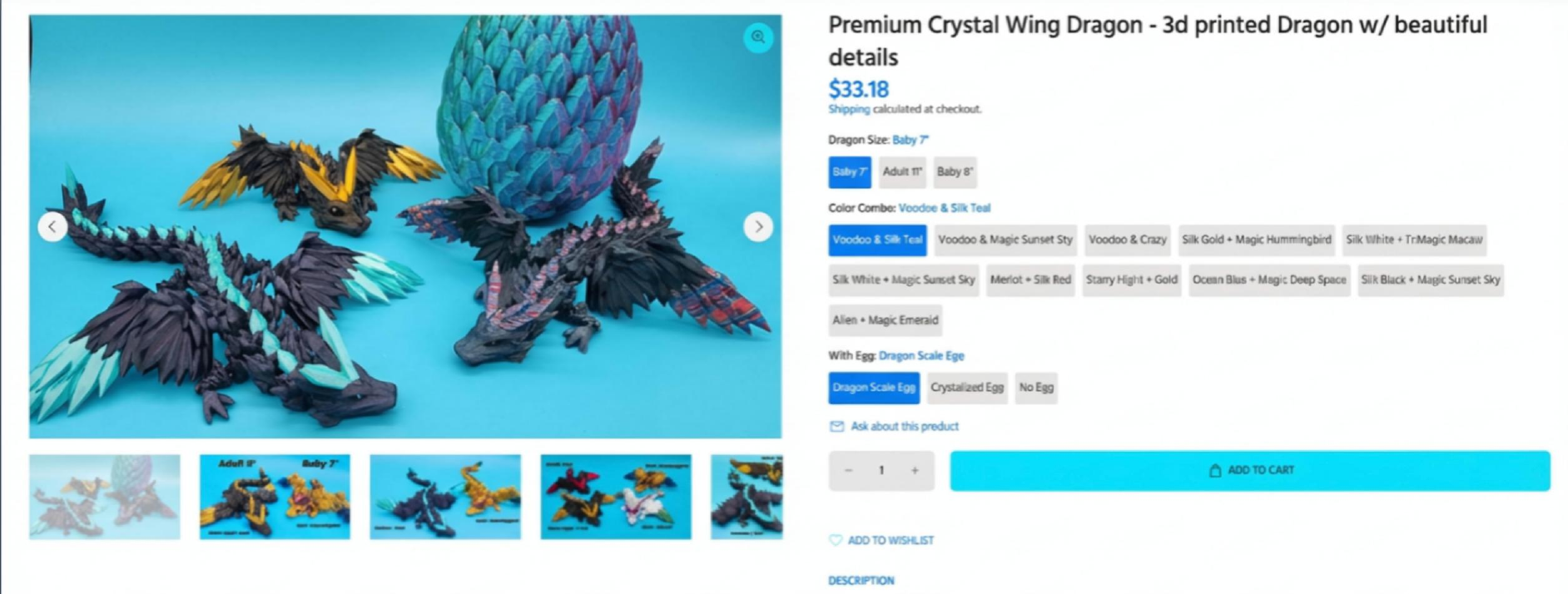}
  \centerline{\small (a) Premium Crystal Wing Dragon product page (\$33.18).}
  
  \vspace{0.5em}
  
  \includegraphics[width=0.9\columnwidth]{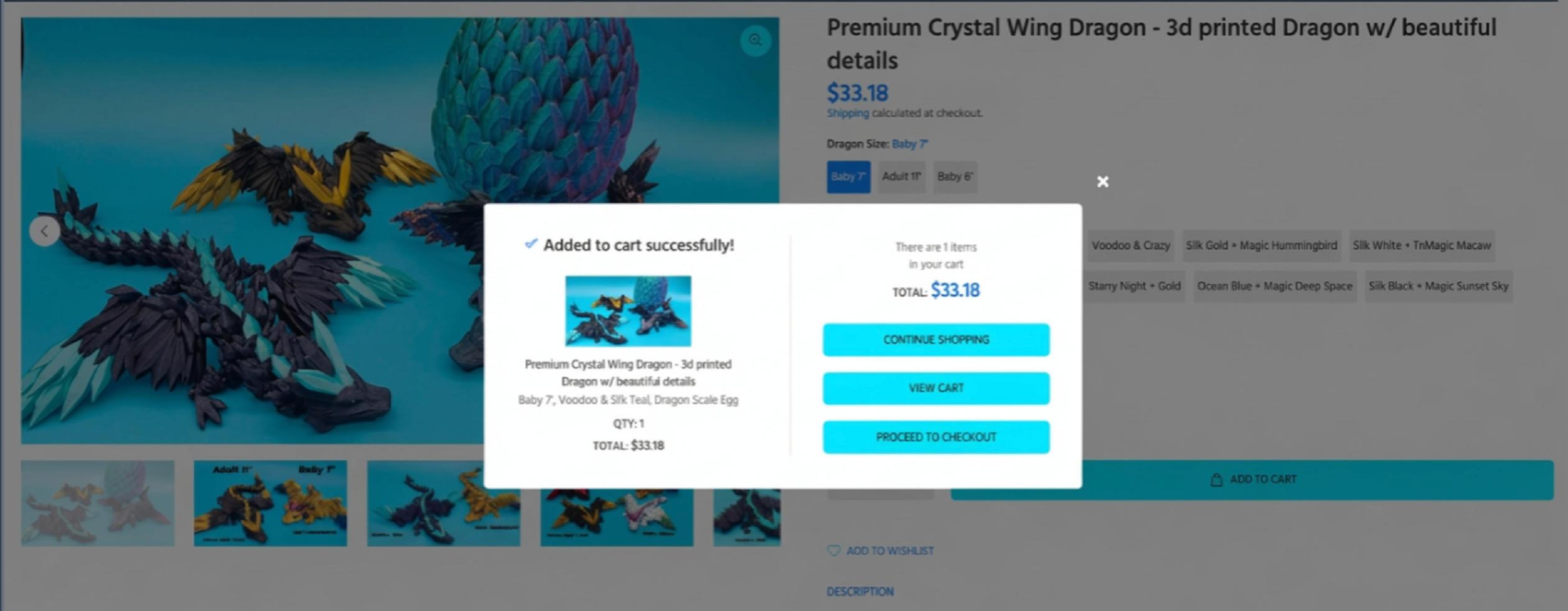}
  \centerline{\small (b) Add-to-cart confirmation.}
  
  \vspace{0.5em}
  
  \captionof{figure}{Product Selection and Purchase.}
  \label{fig:product-cart}
  
\end{minipage}
\end{center}
\begin{center}
  \includegraphics[width=\columnwidth]{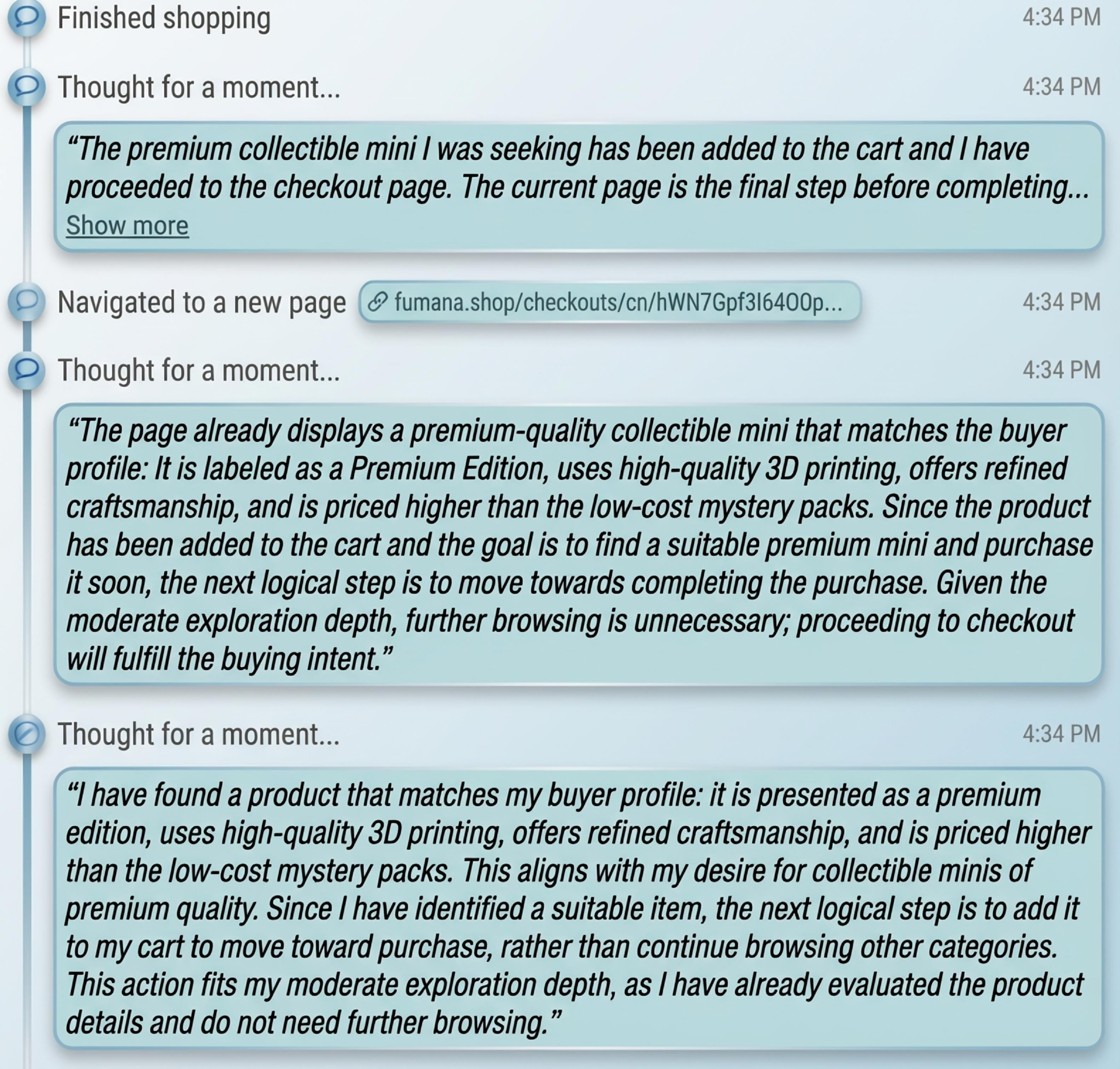}
  \captionof{figure}{Agent Reasoning During Purchase Decision and Checkout.}
  \label{fig:trace-checkout}
\end{center}

\end{multicols}
\end{document}